%% file: paper.tex
\documentclass[10pt,twocolumn,letterpaper]{article}

\usepackage{cvpr}
\usepackage{times}
\usepackage{epsfig}
\usepackage{graphicx}
\usepackage{amsmath}
\usepackage{amssymb}
\usepackage{nicefrac}
\usepackage{fixltx2e} %
\usepackage{booktabs}
\usepackage{leftidx}
\usepackage{appendix}
\usepackage{multirow}
\usepackage{placeins}

\newcommand{\finalcopy}{\cvprfinalcopy}

\graphicspath{{fig/}} %
\input{tables/hpdefs}

\input{plots}

\usepackage[pagebackref=true,breaklinks=true,letterpaper=true,colorlinks,bookmarks=false]{hyperref}
\cvprfinalcopy %
\def\cvprPaperID{4755} %

\ifcvprfinal\fi

\begin{document}

\title{Explicit Spatial Encoding for Deep Local Descriptors}

\author{
Arun Mukundan \qquad  Giorgos Tolias \qquad Ond{\v r}ej Chum\\
{Visual Recognition Group, FEE, CTU in Prague}\\
}

\include{abbrev}

\maketitle

\begin{abstract}

We propose a kernelized deep local-patch descriptor based on efficient match kernels
of neural network activations. Response of each receptive field is encoded together with its
spatial location using explicit feature maps. Two location parametrizations, Cartesian and polar,
are used to provide robustness to a different types of canonical patch misalignment.
Additionally, we analyze how the conventional architecture, \ie a fully connected layer attached after the convolutional part, encodes responses in a spatially variant way.
In contrary, explicit spatial encoding is used in our descriptor, whose potential applications are not limited to local-patches.
We evaluate the descriptor on standard benchmarks. Both versions, encoding $32\!\times\!32$ or
$64\!\times\!64$ patches, consistently outperform all other methods on all benchmarks. The number
of parameters of the model is independent of the input patch resolution.
\end{abstract}

\input{intro}

\input{related}
\input{method}

\input{implementation}

\input{experiments}
\section{Conclusions}
We interpret conventional convolutional local descriptors as efficient match kernels and
show that they learn spatially variant encoding through that last FC layer.
We design a novel local descriptor that explicitly encodes the spatial information.
We use a combined position parametrization handling different sources of geometric misalignment.
It achieves the same performance as state-of-the-art descriptors with fewer parameters and consistently
outperforms them on all standard patch benchmarks with the same number of parameters.

\head{Acknowledgments}
This work was supported by the GA\v{C}R grant 19-23165S, the OP VVV funded project CZ.02.1.01/0.0/0.0/16\_019/0000765 ``Research Center for Informatics'' and the CTU student grant SGS17/185/OHK3/3T/13.

{\small
\bibliographystyle{ieee}
\bibliography{egbib}
}
\end{document}

%% file: tables/hpdefs.tex
\newcommand{\ttb}[1]{{\scshape #1}\xspace}

\newcommand{\les}{{ \hspace{-1pt},\hspace{-1pt}s\hspace{-1pt}=\hspace{0pt} }\xspace}

\newcommand{\indepa}{{\ttb{$\leftidx{^\star}\vpsi^{c}\les1$}}\xspace}
\newcommand{\indepb}{{\ttb{$\leftidx{^\star}\vpsi^{c}\les2$}}\xspace}

\newcommand{\dessum}{\ttb{$\vpsi^{\text{sum}}$}\xspace}
\newcommand{\descat}{\ttb{$\vpsi^{\text{cat}}$}\xspace}

\definecolor{colnsum}{rgb}{0.27344,0.50781,0.70312}%
\definecolor{colncat}{rgb}{0.17969,0.54297,0.33984}%
\definecolor{colours}{rgb}{0.00000,0.53125,0.21484}%
\definecolor{colpola}{rgb}{0.36719,0.23438,0.59766}%
\definecolor{colpolb}{rgb}{0.78906,0.00000,0.12500}%
\definecolor{colcrta}{rgb}{0.97917,0.50000,0.44531}%
\definecolor{colcrtb}{rgb}{0.62500,0.32031,0.17578}%
\definecolor{colinda}{rgb}{0.86719,0.71875,0.52734}%
\definecolor{colhard}{rgb}{0.82031,0.41016,0.11719}%
\definecolor{colindb}{rgb}{1.00000,0.05000,0.47656}%
\definecolor{colnind}{rgb}{0.52734,0.80469,0.91797}%
\definecolor{colhardr}{rgb}{0.12031,0.81016,0.11719}%
\definecolor{colindb64}{rgb}{0.0,0.0,0.0}%
\definecolor{colhard64}{rgb}{0.3,0.5,0.8}%

\definecolor{mycolor1}{rgb}{0.82031,0.41016,0.11719}%
\definecolor{mycolor2}{rgb}{0.00000,0.53125,0.21484}%
\definecolor{mycolor3}{rgb}{0.36719,0.23438,0.59766}%
\definecolor{mycolor4}{rgb}{0.78906,0.00000,0.12500}%
\definecolor{mycolor5}{rgb}{1.00000,0.62500,0.47656}%
\definecolor{mycolor6}{rgb}{0.62500,0.32031,0.17578}%
\definecolor{mycolor7}{rgb}{0.86719,0.71875,0.52734}%
\definecolor{mycolor8}{rgb}{0.97917,0.50000,0.44531}%
\definecolor{mycolor9}{rgb}{0.17969,0.54297,0.33984}%
\definecolor{mycolor10}{rgb}{0.27344,0.50781,0.70312}%
\definecolor{mycolor11}{rgb}{0.52734,0.80469,0.91797}%
\definecolor{mycolor12}{rgb}{0.00000,0.54297,0.54297}%
\definecolor{mycolor13}{rgb}{0.43750,0.50000,0.56250}%
\definecolor{mycolor14}{rgb}{0.43750,0.50000,0.56250}%
\definecolor{mycolor15}{rgb}{0.43750,0.50000,0.56250}%
\definecolor{mycolor16}{rgb}{0.43750,0.50000,0.56250}%
\definecolor{mycolor17}{rgb}{0.43750,0.50000,0.56250}%
\definecolor{mycolor18}{rgb}{0.43750,0.50000,0.56250}%
\definecolor{mycolor19}{rgb}{0.43750,0.50000,0.56250}%
\definecolor{mycolor20}{rgb}{0.43750,0.50000,0.56250}%

\newlength\figH
\newlength\figW

%% file: plots.tex
\usepackage{tikz}
\usetikzlibrary{arrows,shapes,calc,matrix,fit,backgrounds}
\usepackage{pgfplots}
\usepackage{pgfplotstable}
\pgfplotsset{compat=1.9}

\usepackage{xstring}
\usetikzlibrary{patterns}

\usepgfplotslibrary{external}
\IfBeginWith*{\jobname}{fig/extern/}{\finalcopy}{}

\tikzset{every mark/.append style={solid}}
\pgfplotsset{%
	grid=both, width=\columnwidth, try min ticks=5,
	every axis/.append style={font=\scriptsize},
	every axis plot/.append style={thick,mark=none,mark size=1.2,tension=0.18},
	legend cell align=left, legend style={fill opacity=0.8},
}

\pgfplotsset{
	dash/.style={mark=o,dashed,opacity=0.7},
	dott/.style={mark=o,dotted,opacity=0.7},
}

%% file: abbrev.tex
\def\l2{\ensuremath{\ell_2}\xspace}
\def\otimest{\hspace{-3pt}\otimes\hspace{-3pt}}
\def\cdott{\hspace{-2pt}\cdot\hspace{-2pt}}
\def\minust{\hspace{-2pt}-\hspace{-2pt}}
\def\pmm{\ensuremath{\pm}\xspace}

\newcommand{\head}[1]{{\smallskip\noindent\bf #1}}
\newcommand{\alert}[1]{{\color{red}{#1}}}
\newcommand{\eq}[1]{(\ref{eq:#1})\xspace}

\newcommand{\red}[1]{{\color{red}{#1}}}
\newcommand{\blue}[1]{{\color{blue}{#1}}}
\newcommand{\green}[1]{{\color{green}{#1}}}
\newcommand{\gray}[1]{{\color{gray}{#1}}}

\newcommand{\tran}{^\top}
\newcommand{\mtran}{^{-\top}}
\newcommand{\zcol}{\mathbf{0}}
\newcommand{\zrow}{\zcol\tran}

\newcommand{\ind}{\mathbbm{1}}
\newcommand{\expect}{\mathbb{E}}
\newcommand{\nat}{\mathbb{N}}
\newcommand{\zahl}{\mathbb{Z}}
\newcommand{\real}{\mathbb{R}}
\newcommand{\proj}{\mathbb{P}}
\newcommand{\prob}{\mathbf{Pr}}

\newcommand{\mif}{\textrm{if }}
\newcommand{\minimize}{\textrm{minimize }}
\newcommand{\maximize}{\textrm{maximize }}
\newcommand{\st}{\textrm{subject to }}

\newcommand{\id}{\operatorname{id}}
\newcommand{\const}{\operatorname{const}}
\newcommand{\sgn}{\operatorname{sgn}}
\newcommand{\var}{\operatorname{Var}}
\newcommand{\mean}{\operatorname{mean}}
\newcommand{\trace}{\operatorname{tr}}
\newcommand{\diag}{\operatorname{diag}}
\newcommand{\vect}{\operatorname{vec}}
\newcommand{\cov}{\operatorname{cov}}

\newcommand{\softmax}{\operatorname{softmax}}
\newcommand{\clip}{\operatorname{clip}}

\newcommand{\defn}{\mathrel{:=}}
\newcommand{\peq}{\mathrel{+\!=}}
\newcommand{\meq}{\mathrel{-\!=}}

\newcommand{\floor}[1]{\left\lfloor{#1}\right\rfloor}
\newcommand{\ceil}[1]{\left\lceil{#1}\right\rceil}
\newcommand{\inner}[1]{\left\langle{#1}\right\rangle}
\newcommand{\norm}[1]{\left\|{#1}\right\|}
\newcommand{\frob}[1]{\norm{#1}_F}
\newcommand{\card}[1]{\left|{#1}\right|\xspace}
\newcommand{\diff}{\mathrm{d}}
\newcommand{\der}[3][]{\frac{d^{#1}#2}{d#3^{#1}}}
\newcommand{\pder}[3][]{\frac{\partial^{#1}{#2}}{\partial{#3^{#1}}}}
\newcommand{\ipder}[3][]{\partial^{#1}{#2}/\partial{#3^{#1}}}
\newcommand{\dder}[3]{\frac{\partial^2{#1}}{\partial{#2}\partial{#3}}}

\newcommand{\wb}[1]{\overline{#1}}
\newcommand{\wt}[1]{\widetilde{#1}}

\def\xssp{\hspace{1pt}}
\def\ssp{\hspace{3pt}}
\def\msp{\hspace{5pt}}
\def\lsp{\hspace{12pt}}

\newcommand{\cA}{\mathcal{A}}
\newcommand{\cB}{\mathcal{B}}
\newcommand{\cC}{\mathcal{C}}
\newcommand{\cD}{\mathcal{D}}
\newcommand{\cE}{\mathcal{E}}
\newcommand{\cF}{\mathcal{F}}
\newcommand{\cG}{\mathcal{G}}
\newcommand{\cH}{\mathcal{H}}
\newcommand{\cI}{\mathcal{I}}
\newcommand{\cJ}{\mathcal{J}}
\newcommand{\cK}{\mathcal{K}}
\newcommand{\cL}{\mathcal{L}}
\newcommand{\cM}{\mathcal{M}}
\newcommand{\cN}{\mathcal{N}}
\newcommand{\cO}{\mathcal{O}}
\newcommand{\cP}{\mathcal{P}}
\newcommand{\cQ}{\mathcal{Q}}
\newcommand{\cR}{\mathcal{R}}
\newcommand{\cS}{\mathcal{S}}
\newcommand{\cT}{\mathcal{T}}
\newcommand{\cU}{\mathcal{U}}
\newcommand{\cV}{\mathcal{V}}
\newcommand{\cW}{\mathcal{W}}
\newcommand{\cX}{\mathcal{X}}
\newcommand{\cY}{\mathcal{Y}}
\newcommand{\cZ}{\mathcal{Z}}

\newcommand{\vA}{\mathbf{A}}
\newcommand{\vB}{\mathbf{B}}
\newcommand{\vC}{\mathbf{C}}
\newcommand{\vD}{\mathbf{D}}
\newcommand{\vE}{\mathbf{E}}
\newcommand{\vF}{\mathbf{F}}
\newcommand{\vG}{\mathbf{G}}
\newcommand{\vH}{\mathbf{H}}
\newcommand{\vI}{\mathbf{I}}
\newcommand{\vJ}{\mathbf{J}}
\newcommand{\vK}{\mathbf{K}}
\newcommand{\vL}{\mathbf{L}}
\newcommand{\vM}{\mathbf{M}}
\newcommand{\vN}{\mathbf{N}}
\newcommand{\vO}{\mathbf{O}}
\newcommand{\vP}{\mathbf{P}}
\newcommand{\vQ}{\mathbf{Q}}
\newcommand{\vR}{\mathbf{R}}
\newcommand{\vS}{\mathbf{S}}
\newcommand{\vT}{\mathbf{T}}
\newcommand{\vU}{\mathbf{U}}
\newcommand{\vV}{\mathbf{V}}
\newcommand{\vW}{\mathbf{W}}
\newcommand{\vX}{\mathbf{X}}
\newcommand{\vY}{\mathbf{Y}}
\newcommand{\vZ}{\mathbf{Z}}

\newcommand{\va}{\mathbf{a}}
\newcommand{\vb}{\mathbf{b}}
\newcommand{\vc}{\mathbf{c}}
\newcommand{\vd}{\mathbf{d}}
\newcommand{\ve}{\mathbf{e}}
\newcommand{\vf}{\mathbf{f}}
\newcommand{\vg}{\mathbf{g}}
\newcommand{\vh}{\mathbf{h}}
\newcommand{\vi}{\mathbf{i}}
\newcommand{\vj}{\mathbf{j}}
\newcommand{\vk}{\mathbf{k}}
\newcommand{\vl}{\mathbf{l}}
\newcommand{\vm}{\mathbf{m}}
\newcommand{\vn}{\mathbf{n}}
\newcommand{\vo}{\mathbf{o}}
\newcommand{\vp}{\mathbf{p}}
\newcommand{\vq}{\mathbf{q}}
\newcommand{\vr}{\mathbf{r}}
\newcommand{\Vs}{\mathbf{s}}
\newcommand{\vt}{\mathbf{t}}
\newcommand{\vu}{\mathbf{u}}
\newcommand{\vv}{\mathbf{v}}
\newcommand{\vw}{\mathbf{w}}
\newcommand{\vx}{\mathbf{x}}
\newcommand{\vy}{\mathbf{y}}
\newcommand{\vz}{\mathbf{z}}

\newcommand{\vone}{\mathbf{1}}
\newcommand{\vzero}{\mathbf{0}}

\newcommand{\valpha}{{\boldsymbol{\alpha}}}
\newcommand{\vbeta}{{\boldsymbol{\beta}}}
\newcommand{\vgamma}{{\boldsymbol{\gamma}}}
\newcommand{\vdelta}{{\boldsymbol{\delta}}}
\newcommand{\vepsilon}{{\boldsymbol{\epsilon}}}
\newcommand{\vzeta}{{\boldsymbol{\zeta}}}
\newcommand{\veta}{{\boldsymbol{\eta}}}
\newcommand{\vtheta}{{\boldsymbol{\theta}}}
\newcommand{\viota}{{\boldsymbol{\iota}}}
\newcommand{\vkappa}{{\boldsymbol{\kappa}}}
\newcommand{\vlambda}{{\boldsymbol{\lambda}}}
\newcommand{\vmu}{{\boldsymbol{\mu}}}
\newcommand{\vnu}{{\boldsymbol{\nu}}}
\newcommand{\vxi}{{\boldsymbol{\xi}}}
\newcommand{\vomikron}{{\boldsymbol{\omikron}}}
\newcommand{\vpi}{{\boldsymbol{\pi}}}
\newcommand{\vrho}{{\boldsymbol{\rho}}}
\newcommand{\vsigma}{{\boldsymbol{\sigma}}}
\newcommand{\vtau}{{\boldsymbol{\tau}}}
\newcommand{\vupsilon}{{\boldsymbol{\upsilon}}}
\newcommand{\vphi}{{\boldsymbol{\phi}}}
\newcommand{\vchi}{{\boldsymbol{\chi}}}
\newcommand{\vpsi}{{\boldsymbol{\psi}}}
\newcommand{\vomega}{{\boldsymbol{\omega}}}

\newcommand{\rLambda}{\mathrm{\Lambda}}
\newcommand{\rSigma}{\mathrm{\Sigma}}

\def\nnsp{\hspace{-5pt}}
\def\nsp{\hspace{-1pt}}
\def\tsp{\hspace{1pt}}
\def\sssp{\hspace{2pt}}
\def\ssp{\hspace{3pt}}
\def\msp{\hspace{5pt}}
\def\bsp{\hspace{7pt}}
\def\lsp{\hspace{10pt}}

\makeatletter
\DeclareRobustCommand\onedot{\futurelet\@let@token\@onedot}
\def\@onedot{\ifx\@let@token.\else.\null\fi\xspace}
\def\eg{\emph{e.g}\onedot} \def\Eg{\emph{E.g}\onedot}
\def\ie{\emph{i.e}\onedot} \def\Ie{\emph{I.e}\onedot}
\def\cf{\emph{c.f}\onedot} \def\Cf{\emph{C.f}\onedot}
\def\etc{\emph{etc}\onedot} \def\vs{\emph{vs}\onedot}
\def\wrt{w.r.t\onedot} \def\dof{d.o.f\onedot}
\def\etal{\emph{et al}\onedot}
\makeatother

%% file: intro.tex
\section{Introduction}
\label{sec:intro}

Local feature extraction and representation is still an essential part of a number computer vision applications across many different problems. A common and well performing procedure is a sequence of three steps: local feature detection~\cite{HS88,MS04,MTSZMSKG05,BETV09,L04}, local patch rectification into a canonical form, and finally a descriptor construction from the canonical patch~\cite{MS05,BETV09,L04,DS15}. The desired property of the local patch descriptors is that Euclidean distance or a dot product between two descriptors indicates whether they are matching, \ie the local features are coming approximately from the same surface of a 3D scene. The descriptor methods have shifted from hand-crafted to currently the most successful convolutional neural network (CNN) based approaches~\cite{ZK15,STFK+15,BRPM16,MMR+17,TW17,NAS+17}.

Fully convolutional neural networks takes an image or a patch as input and produces a tensor, where a vector at each spatial location can be seen as a detector response over its receptive field. In the case of variable-sized, or non-aligned input, such as images, the response tensor is transformed into a descriptor typically by some form of global pooling~\cite{RSAC14,KMO15,RTC18}, which discards geometric information. The global pooling is analogous to bags-of-features~\cite{CDFWB04,SZ03} or descriptor aggregation~\cite{PSM10,JPDSPS12}. In the case of aligned input of  fixed size, such as rectified image patches, the tensor is vectorized and further processed. Vectorization has similar interpretation to vectorizing spatial bins in SIFT~\cite{L04}. Commonly,  the vectorized tensor is processed by a single fully-connected (FC) layer~\cite{MMR+17,TW17}, that can be either interpreted as learned affine (linear and bias) transformation of the space, \eg whitening and dimensionality reduction, or as spatially dependent embedding with efficient match kernels (EMK)~\cite{BS09,BTJ15} (see Section~\ref{sec:matchkernel}). 
The key contribution of this work is a CNN module that explicitly models the spatial information of a rectified patch.
Its applicability is not limited to local descriptors.

Two rectified patches coming from matching local features are far from being identical in general. The difference has two sources, namely appearance change in imaging process and geometric misalignment. The former comes from different light conditions, non-planarity of the surface, imaging artifacts, \etc. The latter is caused by the detected feature covering slightly different area of the 3D surface, or incorrect rectification of the patch. These are consequences of either the appearance changes or of insufficient geometric invariance of the detector, \ie affine invariant detector acting on projectively transformed surface.

Prior work on hand-crafted feature descriptors has shown that it is beneficial to explicitly address the geometric misalignment. Some of the approaches handling this are soft assignment of gradients to bins in SIFT and continuous spatial encoding by kernel methods in different~\cite{BTJ15} or multiple~\cite{MTB+19} coordinate systems.

CNNs are powerful in modeling the appearance variance, while weak in modeling the geometric displacement (at least with a single FC layer). 
Recent methods propose different ways of incorporating spatial information in a CNN~\cite{NAL+18,LLM+18}, but their application field is different than local descriptors.
In this work, we propose to model the geometric misalignment by efficient match kernels that explicitly encode the spatial positions of the responses. To encode the spatial information, kernel-based explicit feature maps are used in a similar fashion to hand-crafted features~\cite{BTJ15,MTB+19}. 
This can be seen as a transition from soft binning, \ie overlapping receptive fields, to continuous efficient match kernels.
In contrast to models with an FC layer, with efficient match kernels the number of model parameters does not grow with increased resolution of the input patch, \ie the models for $32 \times 32$ patch input has the same number of parameters as the model for $64 \times 64$.
The applications of the proposed descriptor go beyond that of local-patches, \eg tasks where encoding spatial position is essential~\cite{LLM+18,NAL+18}.

The rest of the paper is organized as follows. The related work is discussed in Section~\ref{sec:related}. 
Conventional deep local descriptors and the proposed ones is discussed in Section~\ref{sec:method}. 
Implementation details are detailed in Section~\ref{sec:implementation}.
Finally, we present and discuss our experiments on standard benchmarks in Section~\ref{sec:exp}.

%% file: related.tex
\section{Related work}
\label{sec:related}
In this section, we review prior work related to hand-crafted and learned descriptors of local features.

\subsection{Hand-crafted descriptors} 
There are numerous approaches to hand-craft local descriptors.
The variants are based on different types of processing of the input patch, such as filter-bank responses~\cite{BETV09,BSW05,KY08,OT01,SMo97}, pixel gradients~\cite{L04,MS05,TLF10,AY11}, pixel intensities~\cite{SI07,CLSF10,LCS11,RRKB11} and ordering or ranking of pixel intensities~\cite{OPM02,HPS09}.
The most prominent direction has been that of gradient histograms, an approach followed also by the most popular hand-crafted local descriptor, namely SIFT~\cite{L04}. Several improvements and extensions exist in the literature~\cite{KeS04,YM09,SGS10,SAS07,AZ12,DS15}.
The RootSIFT~\cite{AZ12} variant efficiently estimates Hellinger distance and became a standard choice in approaches and tasks. 

Kernel descriptors are derived from the concept of efficient match kernels~\cite{BS09} and form a flexible way to design descriptors with the desired invariant properties. 
Kernel descriptors have been proposed not only for local patches~\cite{BTJ15} but also as a global image descriptor~\cite{BRF10,BLR+11}. 
The kernel descriptor of Bursuc \etal~\cite{BTJ15} was shown to outperform learned descriptors at that time.

\subsection{Learning  descriptors} 
Structure-from-Motion and datasets such as PhotoTourism~\cite{WB07} gave rise to learned local descriptors.
The learned part varies from their pooling regions~\cite{WB07,SVZ13} and filter banks~\cite{WB07} to transformations for
dimensionality reduction~\cite{SVZ13} and embeddings~\cite{PISZ10}.

Learning is also applied to kernelized descriptors as in the supervised framework by Wang \etal~\cite{WWZX+13}.
The local descriptors in their case are not used separately but directly aggregated into a global image representation, while supervision comes at image level. Kernel local descriptors are combined with supervised learning in the form of discriminative projections in the work of Mukundan \etal~\cite{MTB+19}. 
Our work is inspired by theirs; we use the same kernel-based position encoding, but on top of convoltutional activations instead of pixel attributes.

\subsection{Deep learning of descriptors} 
The interest in local descriptor learning is lately dominated by deep learning~\cite{STFK+15,ZK15,HLJS+15,YTL+16,BRPM16,BJTM16}. 
All examples in the literature use architectures that consist of a sequence of common CNN layers, similar to the ones of generic computer vision tasks, such as object recognition, but less deep and with fewer parameters in total.
They typically require a large amount of training data in the form of local patch pairs or triplets.
Some of the contributions are about mining hard training samples~\cite{STFK+15,MMR+17,LSZ+18}, different loss functions~\cite{BRPM16}, different architectures~\cite{TW17} or training jointly with the local feature detector~\cite{YTL+16}.

Two of the most recent and successful deep local descriptors are L2-Net~\cite{TW17} and HardNet~\cite{MMR+17}.
L2-Net applies the loss function to intermediate feature maps too and the loss function integrates multiple attributes.
HardNet extends L2-Net by sampling the hardest within batch samples and currently constitutes the state-of-the-art descriptor.
Their common characteristic, which is shared among all ancestors, is that they are using common CNN layers in their architecture. 
As a consequence, spatial information of convolutional feature maps is not explicitly encoded, but only processed with a standard FC layer.

%% file: method.tex
\section{Method}
\label{sec:method}

\begin{figure*}[t]
\vspace{15pt}
\centering
\begin{tabular}{c}
\includegraphics[width=0.99\textwidth]{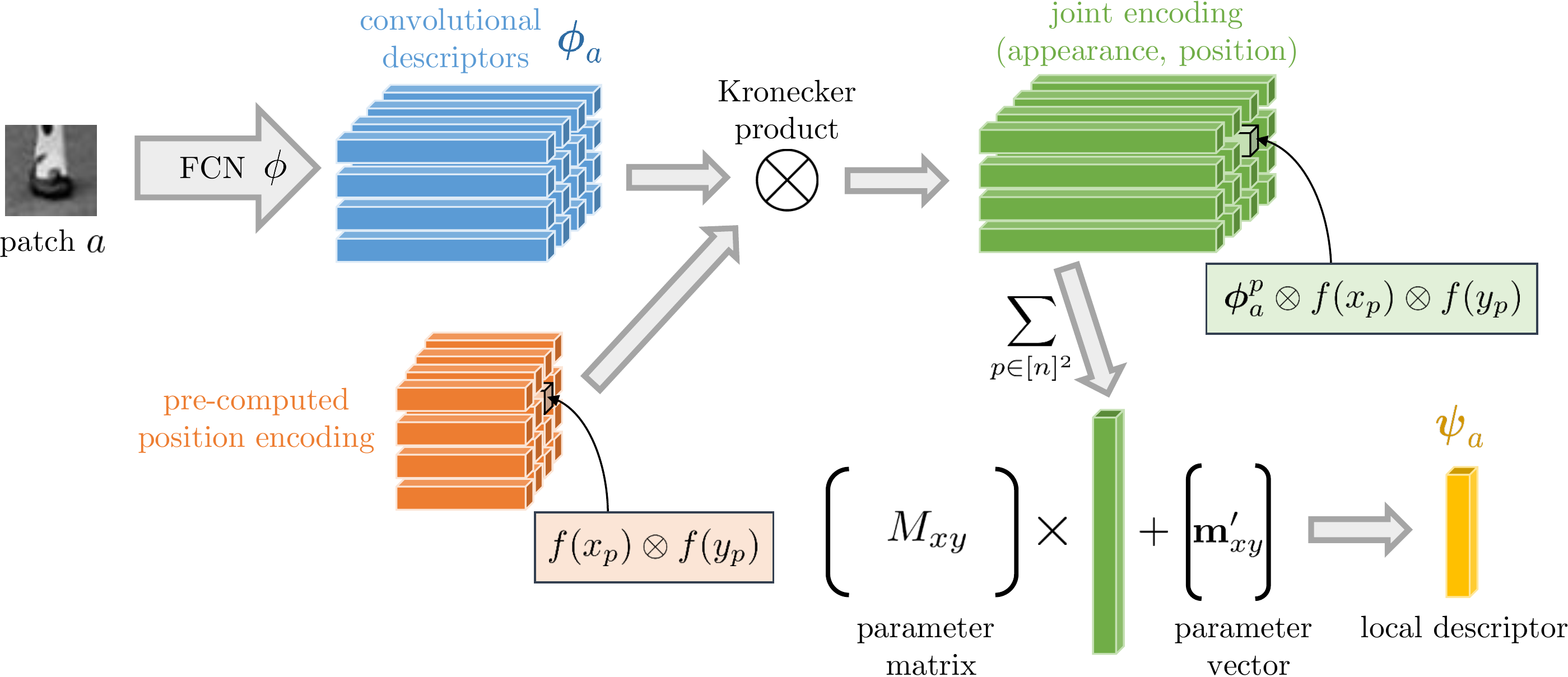}
\end{tabular}
\vspace{15pt}
\caption{Overview of extraction process for the proposed descriptor. We present the case of $\psi^{xy}$ (\ref{equ:cartdes}), while other variants are performed in a similar way.
$\vm'_{xy} = n^2 \vm_{xy}$.
\label{fig:teaser}
\vspace{10pt}}
\end{figure*}

We initially present the current typical architecture for deep local descriptors in the literature.
Then, we provide a different perspective that formulates such descriptors as match kernels.
It allows us to point out how the encoding of convolutional feature maps is performed in a translation variant way, but without explicitly encoding the spatial information.
Finally, we present our novel deep local descriptor which is derived through the same match kernel framework and improves exactly this drawback.
We get inspired by hand-crafted kernel descriptors to incorporate explicit position encoding into deep networks for local descriptors.
An overview of the proposed descriptor is shown in Figure~\ref{fig:teaser},

\subsection{Deep local descriptors} Conventional architectures for deep local descriptors consist of
a sequence of convolutional layers, producing translation invariant feature maps, and a final
FC layer.
We denote the descriptor extraction process by function $\psi: \real^{N\times N} \rightarrow \real^D$,
where $N$ is the size of the input patch and $D$ the dimensionality of the final descriptor.
Descriptor for patch $a \in \real^{N\times N}$ is given by $\psi(a) \in \real^D$ or equivalently $\vpsi_a$ to simplify the notation.

We denote the convolutional part of the network, \ie a \emph{Fully Convolutional Network} (FCN),  by function $\phi: \real^{N\times N} \rightarrow \real^{n \times n \times d}$.
Size $n$ of the resulting feature map is related to input size $N$ and the architecture of the network.
Feature map $\phi(a)$, equivalently denoted by $\vphi_a$, is a 3D tensor of activations, which we also view as a 2D grid of $d$-dimensional vectors.
We call these vectors \emph{convolutional descriptors} and use $\vphi_a^p$ to denote the vector with coordinates $p=(i,j)$ on the $n \times n$ grid, \ie $p \in [n]^2$~\footnote{$[i]=\{1\ldots i\}$ and $[i]^2=[i] \times [i]$}.
Each convolutional descriptor corresponds to a region of the input patch $a$ that is equal to the receptive field size of the feature map.

The standard practice is to vectorize 3D tensor $\vphi_a$ and feed it to an FC layer with parameters that consist of matrix $W \in \real^{D \times(n\times n\times d)}$ and bias $\vw \in \real^D$.
The final descriptor is constructed as
\begin{equation}
\vspace{4pt}
\vpsi_a = W \vect(\vphi_a) + \vw,
\label{equ:stddes}
\vspace{4pt}
\end{equation}
where $\vect$ denotes tensor vectorization. A local descriptor is typically \l2-normalized, which is equivalently achieved by introducing a normalization factor $\gamma_a= \nicefrac{1}{\sqrt{\vpsi_a^\top \vpsi_a}}$ producing descriptor $\hat{\vpsi}_a = \gamma_a \vpsi_a$.

Similarity (or distance) between patches $a$ and $b$ is estimated with inner product (or Euclidean distance) $\hat{\vpsi}_a^\top \hat{\vpsi}_b$.
The \l2-normalized descriptor is always used to compare patches, but we often use $\vpsi_a$ (and not $\hat{\vpsi}_a$) simply to specify which descriptor variant is used. 
Several deep local descriptors in the recent literature, namely L2Net~\cite{TW17}, HardNet~\cite{MMR+17}, and GeoDesc~\cite{LSZ+18} follow such an architecture and can be formulated in the same way.

\subsection{A match-kernel perspective}
\label{sec:matchkernel}

We provide an alternative, but equivalent, construction of deep local descriptors.
We consider matrix $W$ as a concatenation of $n^2$ matrices, \ie
\begin{equation}
\vspace{4pt}
  W = \left( \begin{array}{c} W_{(1,1)}^\top \\ \vdots \\ W_{(i,j)}^\top \\ \vdots \\ W_{(n,n)}^\top \end{array} \right)^\top,
\vspace{4pt}
\end{equation}
where $W_{p} \in \real^{D \times d}$. Descriptor in (\ref{equ:stddes}) can be now written as
\begin{equation}
\vspace{4pt}
\vpsi_a = \sum_{p \in [n]^2 } W_p \vphi_a^p + \vw',
\label{equ:alterdes}
\vspace{4pt}
\end{equation}
where $\vw' = \vw / n^2$.
Moreover, patch similarity  becomes
\begin{align}
  \hat{\vpsi}_a^\top \hat{\vpsi}_b \propto & \sum_{p, q\in [n]^2} \left(W_p \vphi_a^p + \vw' \right)^\top \left(W_q \vphi_b^q + \vw' \right) \nonumber\\
                                         = & \sum_{p, q\in [n]^2} g_{fc}(\vphi_a^p, p)^\top g_{fc}(\vphi_b^q, q),
\label{equ:stdmker}
\end{align}
where $g_{fc}: \real^d \times [n]^2 \rightarrow \real^D$ is a function that encodes a convolutional descriptor in a translation variant way, depending on its position in the $n \times n$ grid.
The match kernel formulation in (\ref{equ:stdmker}) interprets deep local descriptor similarity as similarity accumulation for all pairs of positions on the $n \times n$ grid. It reveals that matching between convolutional descriptors in $\vphi_a$ and $\vphi_b$ is performed in a translation variant way.
The \emph{encoding function} $g$ in the case of conventional deep local descriptors is 
\begin{equation}
\vspace{4pt}
g_{fc}(\vv, p) = W_p \vv +\vw',
\vspace{4pt}
\end{equation}
where matrix $W_p$ and $\vw'$ come from the parameters of the FC layer.
In this work, we propose a new encoding function $g$, not restricted to standard CNN architecture (layers), that explicitly encodes position $p$ on the 2D grid.

\subsection{Position encoding} 

Explicit feature maps~\cite{VZ10} are used to encode the position. Let $f: \real \rightarrow \real^{2s+1}$ be a feature map, where $s$ is a design choice defining the dimensionality of the embedding. Such a feature map defines a shift invariant kernel $K: \real \times \real \rightarrow \real$ with kernel signature $k$, so that $K(\alpha,\beta) = k(\alpha-\beta)$
\begin{equation}
\vspace{4pt}
f(\alpha)^\top f(\beta) = K(\alpha, \beta) = k(\alpha-\beta).
\label{equ:kerapprox}
\vspace{4pt}
\end{equation}
The kernel $K$ (or the feature map $f$) is constructed to approximate the Von Mises kernel~\cite{TBFJ15}.

We propose encoding function $g_{xy}: \real^d \times [n]^2 \rightarrow \real^{D(2s+1)^2}$ given by
\begin{equation}
\vspace{4pt}
g_{xy}(\vphi_a^p, p) = \vphi_a^p \otimes f(x_p) \otimes f(y_p),
\label{equ:cartenc}
\vspace{4pt}
\end{equation}
where $\otimes$ is the Kronecker product and $x_p$ and $y_p$ provide the coordinates of position $p$ in a Cartesian coordinate system~\footnote{For $p=(i,j)$, $x_p=i$ and $y_p=j$.}.
It is a joint encoding of the convolutional descriptor and the explicit representation of its position.
It is inspired by the work of Mukundan \etal~\cite{MTB+19} who propose a hand-crafted local descriptor that encodes pixel gradients with their positions in the patch.
Similarity of two such encodings is given by
\begin{align}
g_{xy}(\vphi_a^p, p)^\top g_{xy}(\vphi_b^q, q) = 
{\vphi_a^p}^\top \vphi_b^q \cdott k(x_p\minust x_q) \cdott k(y_p\minust y_q).
\label{equ:encsim}
\end{align}
It is equivalent to the product of descriptor similarity and similarity of positions on the Cartesian grid.

Following the paradigm of descriptor whitening of hand-crafted descriptors~\cite{MTB+19,BLVM17}, we propose the final local descriptor 
\begin{align}
\vspace{4pt}
\vpsi_a^{xy} &= \sum_{p \in [n]^2 } w_p M_{xy} g_{xy}(\vphi_a^p, p) +\vm_{xy} \\
						 &= M_{xy} \left( \sum_{p \in [n]^2 } w_p  g_{xy}(\vphi_a^p, p)\right) +n^2 \vm_{xy}, 
\label{equ:cartdes}
\vspace{4pt}
\end{align}
where $M_{xy} \in \real^{D\times d(2s+1)^2}$ and $\vm_{xy} \in \real^D$ are parameters to be learned during training, while $w_p = \exp(-\rho_p^2)$ is a weight giving importance according to the distance $\rho_p$ from the center of the patch.
Note that in contrast to (\ref{equ:alterdes}) the same matrix, \ie $M_{xy}$, is used for all convolutional descriptors.
As a result the number of required parameters is reduced and multiplication by $M_{xy}$ can be efficiently performed after the summation (\ref{equ:cartdes}).
In analogy to the encoding of position in a Cartesian coordinate system, 
we additionally propose the encoding \wrt a polar coordinate system\footnote{For $p=(i,j)$, $\rho_p=\sqrt{(i-c)^2+(j-c)^2}$ and $\theta_p=\tan^{-1} \frac{j-c}{i-c}$, where $c=(n+1)/2$.} by
\begin{equation}
\vspace{4pt}
g_{\rho\theta}(\vphi_a^p, p) = \vphi_a^p \otimes f(\rho_p) \otimes f(\theta_p),
\label{equ:polarenc}
\vspace{4pt}
\end{equation}
and the corresponding descriptor
\begin{equation}
\vspace{4pt}
\vpsi_a^{\rho\theta} = \sum_{p \in [n]^2 } w_p M_{\rho\theta} g_{\rho\theta}(\vphi_a^p, p) +\vm_{\rho\theta}.
\label{equ:polardes}
\vspace{4pt}
\end{equation}
Different parameterizations, \ie using different coordinate system, provide tolerance to different kinds of misalignment between patches. 
Cartesian offers tolerance to translation misalignment, while polar offers tolerance to rotation and scale misalignment. %
To benefit from both types of tolerance, we further use the combined encoding that uses the two coordinate systems and is produced by concatenation of the previous encoding. It is defined as function $g_{c}: \real^d \times \real^d \times [n]^2 \rightarrow \real^{2D(2s+1)^2}$ given by 
\begin{equation}
\vspace{4pt}
\small
g_{c}(\vphi_a^p, \tilde{\vphi}{}_a^p, p) \hspace{-2pt}= \hspace{-3pt}  \left( (\vphi_a^p \otimest f(x_p) \otimest f(y_p))^{\hspace{-2pt}\top}\hspace{-2pt} ,\hspace{-2pt} (\tilde{\vphi}{}_a^p \otimest f(\rho_p) \otimest f(\theta_p))^{\hspace{-2pt}\top}  \right)^{\hspace{-2pt}\top} \hspace{-3pt}
\label{equ:jointenc}
\vspace{4pt}
\end{equation}
where $\tilde{\vphi}$ is used to show that the two encodings do not need to rely on the same FCN $\phi$.
Subscript $c$ refers to the combined coordinate system, but we skip $xy\rho\theta$ to simplify the notation.
The final descriptor proposed in this work is
\begin{equation}
\vspace{4pt}
\leftidx{^\star}\vpsi_a^{c} =  \sum_{p \in [n]^2 } w_p M_{c} g_{c}(\vphi_a^p, \tilde{\vphi}{}_a^{p}, p) +\vm_{c}
\label{equ:des}
\vspace{4pt}
\end{equation}
where $M_{c} \in \real^{D\times 2d(2s+1)^2}$, and left superscript $\star$ is used to denote that a separate FCN is used for each encoding, correspondingly coordinate system.

%% file: implementation.tex
\begin{table*}[t]
\vspace{15pt}
\input{tables/params}

\vspace{20pt}
\caption{Number of parameters for different models.
The convolutional part $\phi$ has identical architecture for all models.
Cases where both $\phi$ and $\tilde{\phi}$ appear use a separate convolutional part for the Cartesian and the polar descriptor.
These specifications correspond to $d=128$, and $D=128$. The resulting $n$ is equal to 8 and 16 for $N$ equal to 32 and 64, respectively.
We report $M$ and $\vm$ due to limited space, but we refer to $M_{xy}$, and $M_{c}$ according to the respective table, and similarly for $\vm$. Descriptor $\vpsi^{\rho\theta}$ has identical requirements as descriptor $\vpsi^{xy}$.
The parameter requirements of our descriptor remain unchanged for different patch size $N$.
\label{tab:params}
\vspace{10pt}}
\end{table*}
\section{Implementation details}
\label{sec:implementation}
In this section, we provide implementation details that concern the efficiency of the aggregation, describe the different architectures and their required number of parameters, and finally discuss the training procedure.

\subsection{Efficient aggregation}
We describe the implementation details for variant $\psi_a^{xy}$, but these hold for other variants too in the same way.
Vectors $ w_p f(x_p) \otimes f(y_p) \in \real^{(2s+1)^2}$ that encode positions $p \in [n]^2$ are fixed for the 2D grid of size $n \times n$.
Thus, we pre-compute and store them in matrix $F \in \real^{n^2 \times (2s+1)^2}$.
We reshape 3D tensor $\vphi_a$ into matrix $\Phi \in \real^{n^2 \times d}$.
Given these two matrices and due to the linearity of matrix to vector multiplication we can re-write the descriptor as
\begin{align}
\vpsi_a^{xy} &= \sum_{p \in [n]^2 } w_p M_{xy} g_{xy}(\vphi_a^p, p) +\vm_{xy},  \nonumber\\
						 &= M_{xy} \left(\sum_{p \in [n]^2 }  w_p g_{xy}(\vphi_a^p, p) \right) +n^2 \vm_{xy},  \label{equ:desnoeffic} \\
						 &= M_{xy} \left(\sum_{p \in [n]^2 }  w_p \vphi_a^p \otimes f(x_p) \otimes f(y_p) \right) +n^2 \vm_{xy}, \nonumber \\
						 &= M_{xy} \vect(\Phi^\top F) +n^2 \vm_{xy}. \label{equ:deseffic}
\end{align}
Multiplication $\Phi^\top F$ makes the computation memory efficient because it avoids explicit storing of the Kronecker product for each $p$. To evaluate (\ref{equ:desnoeffic}), the memory requirements are $n^2d(2s+1)^2$ numbers, while to evaluate (\ref{equ:deseffic}), only 
$n^2 (d+(2s+1)^2)$ numbers are allocated. Using setup $d = 128$ and $s = 2$ in our experiments, the memory requirements are reduced by a factor of $20.9$.

\subsection{Architecture}

We use the HardNet+~\cite{MMR+17} architecture for the convolution part, since HardNet+ achieves state-of-the-art performance on all benchmarks. We also use it a baseline to compare with.

The statistics of the convolutional part $\phi$ are described in Table~\ref{tab:params} (left). Each convolutional layer is
followed by batch normalization and ReLU, while no bias is used. Table~\ref{tab:params} (right) provides the total number of parameters for HardNet+ and our networks, namely, plain polar or Cartesian encoding with different dimensionality of the explicit feature maps ($s=1$ and $s=2$ frequencies used), and the joint encoding with a common ($\phi$) or separate ($\phi$ and $\tilde{\phi}$) convolutional part. Note that for the joint encoding with separate convolutional parts and $s = 2$ frequencies, the proposed network needs roughly the same number of parameters as HardNet+ with input patch of size $32 \times 32$ pixels ($N = 32$). In all other settings of the proposed architecture, the number of parameters is significantly reduced. Importantly, the number of parameters for larger patch sizes (such as $64 \times 64$), that provide better performance, the number of parameters stays fixed for the proposed architecture. For Hardnet+, the number of parameters of the FC layer increases  by a factor of 4 for $64 \times 64$ input patches.

\subsection{Training}

We would like to highlight the contribution of the explicit spatial encoding and to provide direct comparison to the current state-of-the-art descriptor construction. To avoid changing many things at the same time, we follow exactly the same training procedure as HardNet+, which we briefly review below.

The network is trained with the triplet loss defined as 
\begin{equation}
\ell(\hat{\psi}_{an}, \hat{\psi}_{pos}, \hat{\psi}_{neg})= [1-||\hat{\psi}_{an}-\hat{\psi}_{pos}||+||\hat{\psi}_{an}-\hat{\psi}_{neg}||]_{+},
\end{equation}
acting on a triplet formed by an anchor, a positive (matching to the anchor), and a negative (non-matching to the anchor) descriptor.
A batch of size 1024 patches is constructed from 512 pairs of anchor-positive descriptors. 
Regarding a particular pair in the batch, the positive descriptors of all other pairs are considered as candidate negatives.
Finally, the one with the smallest Euclidean distance to the anchor within the batch is chosen as a hard negative to form a triplet.

We use Stochastic Gradient Descent (SGD) to perform the training.
The total training set consists of 2 million anchor-positive pairs and the training lasts 10 epochs.
Data augmentation is employed by random patch rotation, scaling and flipping.
The learning rate is set to 10, and linearly decays to zero withing 10 epochs.
Momentum is equal to 0.9 and weight decay to $10^{-4}$.
Random orthogonal initialization is used for the weights of the network~\cite{SAMJ13}.
The method is implemented in the PyTorch framework. %

%% file: tables/params.tex
\small
\centering
\setlength\extrarowheight{2pt}
\begin{tabular}{ccc}
\multirow{2}{*}{
\begin{tabular}{@{\ssp}c@{\bsp}r@{\bsp}r@{\ssp}}
\multicolumn{3}{c}{Convolutional part $\phi$} \\
\hline
  Conv. layer &  Param. matrix shape & \# Parameters \\
  \hline
  1     &         [ 1,  32, 3, 3 ]   &              288 \\
  2     &        [ 32,  32, 3, 3 ]   &            9,216 \\
  3     &        [ 32,  64, 3, 3 ]   &           18,432 \\
  4     &        [ 64,  64, 3, 3 ]   &           36,864 \\
  5     &        [ 64, 128, 3, 3 ]   &           73,728 \\
  6     &       [ 128, 128, 3, 3 ]   &          147,456 \\
  \hline
  Total &                         &          285,984 \\
  \hline
\end{tabular}
} %
&
\begin{tabular}{@{\lsp}l@{\lsp}r@{\lsp}r@{\lsp}}
&&\\
HardNet &$N=32$ &$N=64$\\ \hline
$\vphi$ &        285,984 & 285,984  \\
FC      &        1,048,576 & 4,194,304\\
\hline
Total   &        1,334,560 & 4,480,288\\
\hline
\end{tabular}
&
\begin{tabular}{@{\lsp}l@{\lsp}r@{\lsp}r@{\lsp}}
$\psi^{xy}$ & \multicolumn{2}{c}{$N\hspace{-3pt}=\hspace{-3pt}\{32,64\}$} \\\hline
& $s\hspace{-3pt}=\hspace{-3pt}1$ & $s\hspace{-3pt}=\hspace{-3pt}2$\\
\cmidrule(lr){2-3}
$\vphi$ &          285,984 & 285,984 \\
$M$, $\vm$  &      147,584 & 409,728 \\
\hline
Total       &      433,568 & 695,712 \\
\hline
\end{tabular}
\\[40pt]
& 
\begin{tabular}{@{\lsp}l@{\lsp}r@{\lsp}r@{\lsp}}
&&\\
$\psi^{c}$ & \multicolumn{2}{c}{$N\hspace{-3pt}=\hspace{-3pt}\{32,64\}$} \\\hline
& $s\hspace{-3pt}=\hspace{-3pt}1$ & $s\hspace{-3pt}=\hspace{-3pt}2$\\
\cmidrule(lr){2-3}
$\vphi$ &          285,984 & 285,984  \\
$M$, $\vm$  &      295,040 & 819,328\\
\hline
Total       &      581,024  & 1,105,312\\
\hline
\end{tabular}
&
\begin{tabular}{@{\lsp}l@{\lsp}r@{\lsp}r@{\lsp}}
$\leftidx{^\star}\psi^{c}$ & \multicolumn{2}{c}{$N\hspace{-3pt}=\hspace{-3pt}\{32,64\}$} \\\hline
& $s\hspace{-3pt}=\hspace{-3pt}1$ & $s\hspace{-3pt}=\hspace{-3pt}2$\\
\cmidrule(lr){2-3}
$\vphi$           &  285,984 & 285,984  \\
$\tilde{\vphi}$   &  285,984 & 285,984 \\
$M$, $\vm$        &  295,040 & 819,328\\
\hline
Total       &      867,008 & 1,391,296\\
\hline
\end{tabular}

\\
\end{tabular}

%% file: experiments.tex
\section{Experiments}
\label{sec:exp}
We first describe the datasets and the evaluation protocols used in our experiments,
and then present qualitative results showing the impact of the training on patch similarity.
Finally we present the results achieved by different variants of our descriptor and show a comparison with the state of the art.

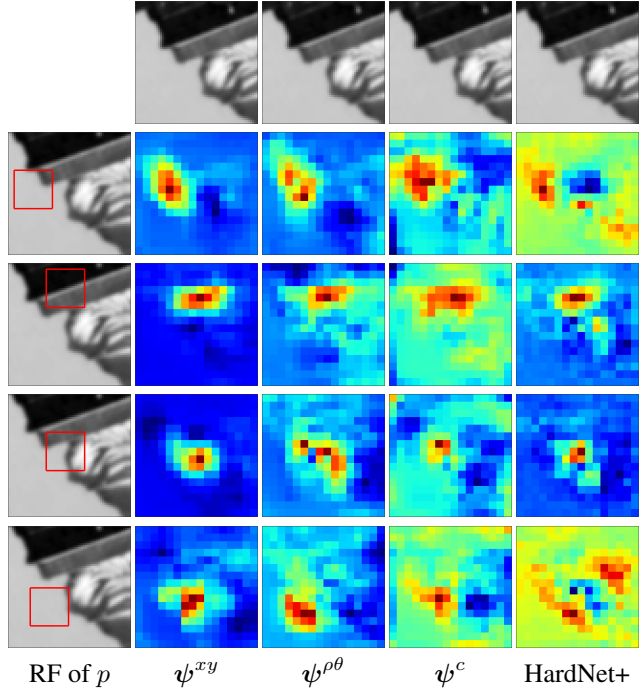
\begin{figure}[t]
  \input{tables/indepnew_patchmaps}
  \vspace{10pt}
  \caption{Visualization of similarity between position $p$ of the $n \times n$ grid (rows) on a patch and another whole patch for different methods (columns). Heat-maps are normalized to $[0, 1]$ with red corresponding to the maximum similarity.
  Red box is used to depict the receptive field (RF) of $p$.\label{fig:maps1}
  \vspace{-10pt}
  }
\end{figure}

\subsection{Datasets and protocols.}
We use two publicly available patch datasets, namely \emph{PhotoTourism} (PT)~\cite{WB07}
and \emph{HPatches} (HP)~\cite{BLVM17}. We use the former for both training and evaluation, while
the latter only for evaluation when training on PT to show the generalization ability of the descriptor.

The PT dataset consists of following 3 separate sets, Liberty, Notredame and Yosemite.
Each consists of local features detected with the Difference-of-Gaussians (DoG) detector
and verified through an SfM pipeline.
Each set comprises about half a million $64 \times 64$ patches, associated
with a discrete label which is the outcome of SfM verification.
The test set consists of 100k pairs of patches corresponding to the same (positive) 3D point,
and an equal number corresponding to different (negative) 3D points.
The metric used to measure performance is the \emph{false positive rate at 95\% of recall} (FPR@95).
Models are trained on one set and tested on the other two, and the mean of 6 scores is reported.

The HP dataset contains patches of higher diversity and is more realistic.
Evaluation is performed on three different tasks, namely verification, retrieval, and matching.
Despite the fact that we do not train on HP, we evaluate on all 3 train/test splits and report the average performance to allow future comparisons.
We follow the common practice and train our descriptor on Liberty of PT to evaluate on HP.

We repeat each experiment three times, with different random seeds to initialize the parameters, and report mean and standard deviation of the 3 runs.
We followed this policy for all variants and datasets.

Recently, larger and more diverse datasets~\cite{MDG+18,LSZ+18} have been introduced to improve local descriptor training. These are shown to improve the performance of state-of-the-art descriptors even by simply replacing the training dataset. We have not included them in our experiments but expect the impact to be similar on our descriptor too.

\begin{table*}[t]
\vspace{20pt}
\input{tables/pt}
\vspace{15pt}
\caption{Performance comparison of the proposed descriptors with the state-of-the-art descriptor HardNet+ on the PhotoTourism dataset.
  Performance is measured via FPR@95.
  We repeat each experiment/training 3 times and report mean performance and standard deviation.
  Patch size is $N = 32$.
  $\dagger$: Reported in the original work.\label{tab:pt}
 \vspace{15pt}
 }
\end{table*}

\subsection{Visualizing patch similarity.}
We construct encodings $g(\vv, p)$, before aggregation, for our descriptors and for the conventional case and construct a similarity map to analyze the impact of the position encoding. We present such visualization in Figure~\ref{fig:maps1}.
We pick position $p$ and compute similarity $g_{fc}(\vphi_a^p, p)^\top g_{fc}(\vphi_b^q, q), \forall q \in [n]^2$, for the conventional case, and $(M_c g_c(\vphi_a^p, p) +\vm_c)^\top (M_c  g_c(\vphi_b^q, q) + \vm_c), \forall q \in [n]^2$ for ours in the case of the combined descriptor. %
We observe how all architectures, including the conventional one, result in large similarity values near $p$.

\subsection{Results and comparisons.}

We train and evaluate different variants of the proposed descriptor. If not otherwise stated, we use input patches of size equal to $32\times 32$, which is the standard practice for deep local descriptors. We further examine the case of $64\times 64$ input patches.
We always set $d=128$ and $D=128$. The dimensionality of the feature maps is controlled by $s$ which we set equal to $1$ or $2$ in our experiments.

\textbf{Reproducing HardNet+.} Our implementation, training procedure, and training hyper-parameters are based on HardNet+~\footnote{\url{https://github.com/DagnyT/hardnet}}.
We reproduce its training and report our own results, proving that our benefit is not an outcome of implementation details.
We report both the achieved performed in the original publication and our reproduced ones in all the comparisons.

\textbf{Baselines for ablation study.} We train and test the following two baselines to see the impact of the position encoding.
First, we train a descriptor that encodes convolutional descriptors in $\vphi_a$ in a translation invariant way, \ie no position encoding at all. It is implemented by spatial sum pooling on $\vphi_a$ and given by

\begin{equation}
\vpsi_a^{\text{sum}} = \sum_{p\in [n]^2} \vpsi_a^{p}.
\label{equ:sumdes}
\end{equation}
The dimensionality of $\vpsi_a^{\text{sum}}$ is equal to $d$ and not $D$ in this case. However, $d=D=128$, making this descriptor directly comparable to all others.

Second, we train a descriptor that encodes the spatial information simply by concatenation, \ie vectorization of $\vphi_a$, which does not provide any tolerance to position misalignments. It is given by
\begin{equation}
\vpsi_a^{\text{cat}} = \vect \vpsi_a
\label{equ:catdes}
\end{equation}
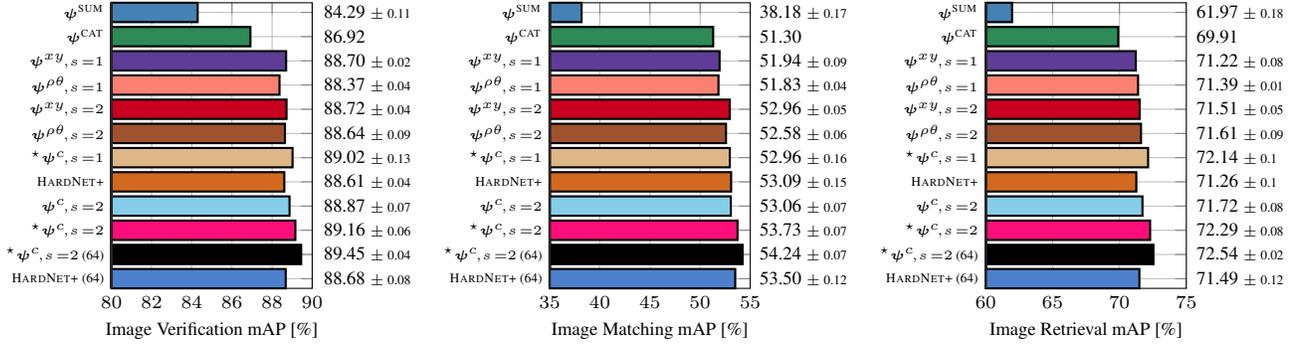
\begin{figure*}
\input{tables/hp_internal}
\caption{Performance comparison on the HPatches benchmark.
The training is performed on the Liberty set of PhotoTourism dataset for all descriptors and with identical setup.
Performance is measured via mean Average Precision (mAP).
We repeat each experiment/training 3 times and report mean performance and standard deviation (with the exception of \descat that due to very high dimensionality was trained only once).
All descriptors have 128 dimensions, with the exception of \descat which has 8192.
The methods are sorted \wrt the required number of parameters (top is the least demanding, \ie less parameters).
All methods are trained and tested with patch size $N=32$ unless when (64) is reported.
\label{fig:hp_internal}
}
\end{figure*}
\begin{figure*}
\input{tables/hp_stateofart}
\caption{Performance comparison with the state of the art on the HPatches benchmark.
The learning for learned descriptors is performed on the Liberty set of PhotoTourism dataset.
Hand-crafted descriptors are shown with striped bars.
Performance is measured via mean Average Precision (mAP).
The performance of our descriptor is the mean of 3 repetitions of each experiment/training.
All methods are trained and tested with patch size $N=32$ unless when (64) is reported.
$\dagger$: Reported in the original work.
\label{fig:hp_internal}
}
\end{figure*}
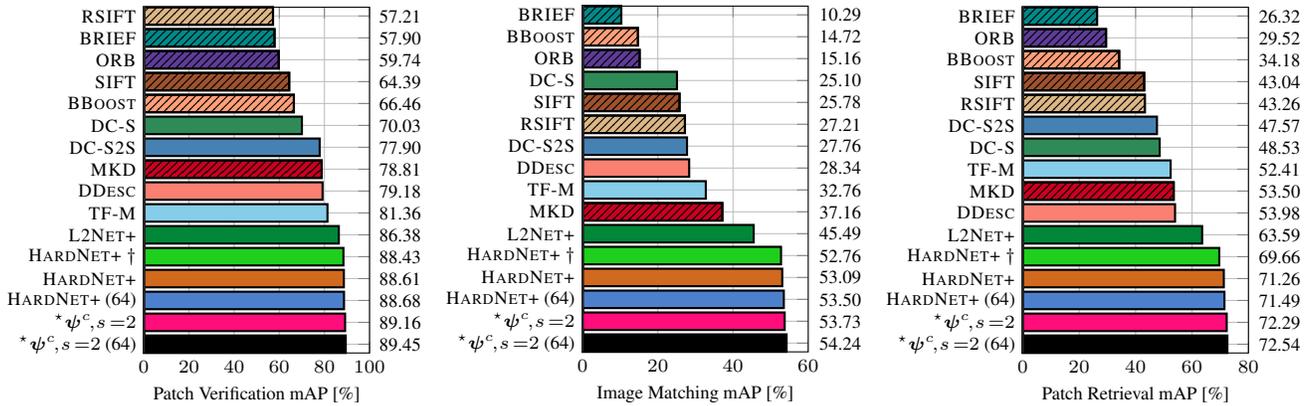

\textbf{Impact of position encoding.}
We compare our descriptor with HardNet+ on PT and show results in Table~\ref{tab:pt}.
Conceptually it is a comparison between the conventional architecture that uses an FC layer to ``feed'' the convolutional descriptors to, and our kernel-based approach to explicitly encode the spatial information.
Our descriptors (with $s=2$) slightly outperforms HardNet+ while it has roughly the same number of parameters.
Even the variant with fewer parameters ($s=1$) performs similarly.

A more thorough comparison, examining the impact of the explicit spatial encoding, is performed on HP and presented in Figure~\ref{fig:hp_internal}.
Firstly, we evaluate \dessum as part of an ablation study.
It is translation invariant that totally discards the spatial information.
It does not require additional parameters other than the ones for FCN $\phi$.
It has significantly lower performance compared to all the other descriptors.
We additionally tried including multiplication by matrix $M_{\text{sum}}$ in (\ref{equ:sumdes}) and did not notice performance improvements.
Descriptor \descat is another case not requiring additional parameters.
It is translation variant in a ``rigid'' way, whose tolerance to translation misalignment is restricted to the amount that the large receptive field offers. Despite the very large dimensionality, it is not a top performer.
Even our light-weight variant with as few as 127k additional parameters (excluding $\phi$) recovers most of the performance loss due to lack of spatial information, \ie \wrt \dessum.
This result suggests that the common choice of an FC layer for deep local descriptors might be over-parametrized.
It is not the best performing either. Our variant \indepb is consistently the top performing one on all tasks.

\textbf{Comparison with the state of the art.} We finally present a comparison to the state of the art on HP in Figure~\ref{fig:hp_internal}.
The comparison includes a set of hand-crafted and learned local descriptors, namely RSIFT~\cite{AZ12}, SIFT~\cite{L04}, BRIEF~\cite{CLSF10}, BBoost~\cite{TCL+12}, ORB~\cite{RRKB11}, MKD~\cite{MTB+19}, DeepCompare~\cite{ZK15}, DDesc~\cite{STFK+15}, TFeat~\cite{BRPM16}, L2Net~\cite{TW17} and HardNet~\cite{MMR+17}.
The proposed descriptor achieves the best performance with a 128D descriptor on all 3 tasks consistently.

%% file: tables/indepnew_patchmaps.tex
\centering
\begin{tabular}{@{\sssp}c@{\sssp}c@{\sssp}c@{\sssp}c@{\sssp}c@{\sssp}}
  & \includegraphics[height=46pt]{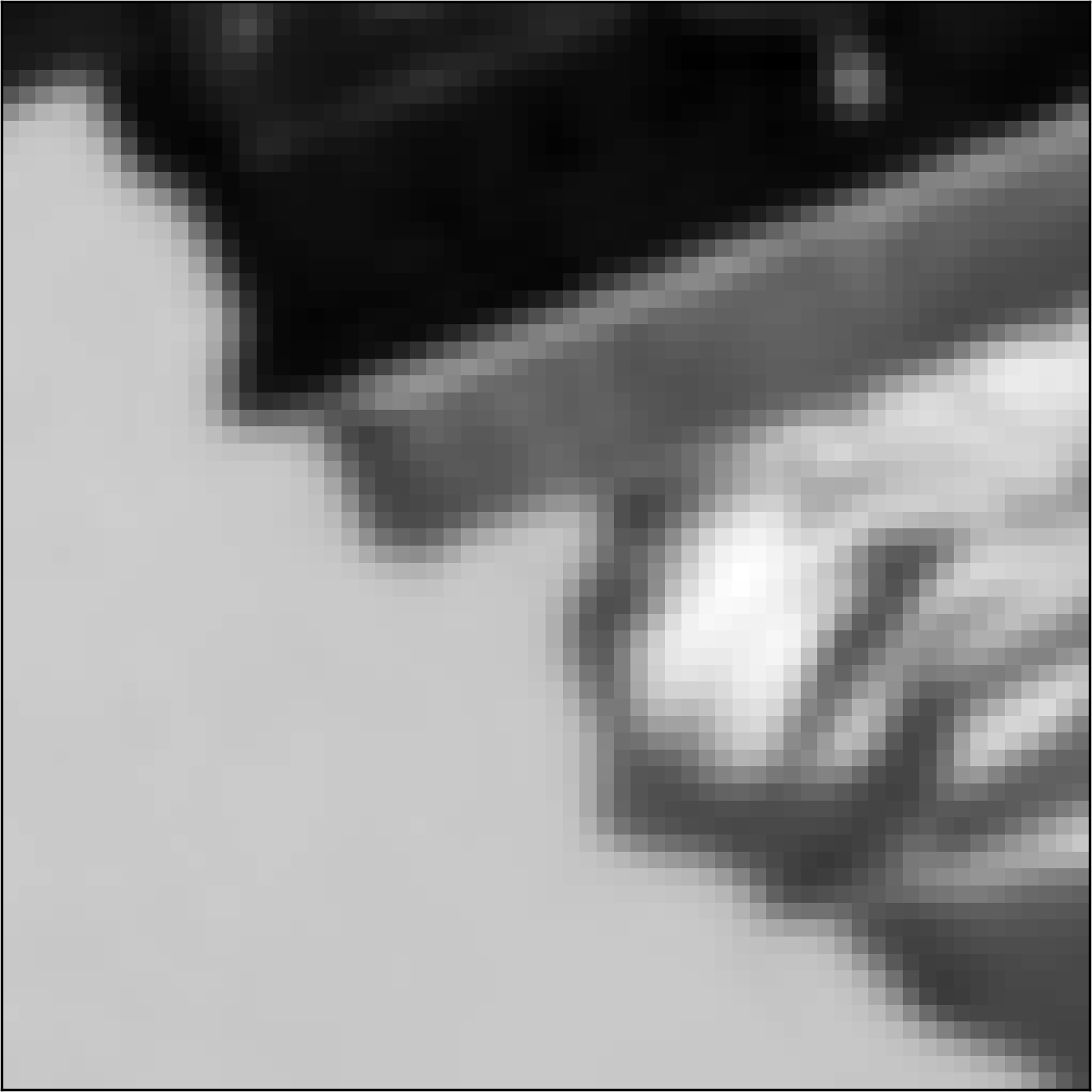}      %
  & \includegraphics[height=46pt]{patchmaps/cart64/patch2_0.png}      %
  & \includegraphics[height=46pt]{patchmaps/cart64/patch2_0.png}      %
  & \includegraphics[height=46pt]{patchmaps/cart64/patch2_0.png} \\   %
\includegraphics[height=46pt]{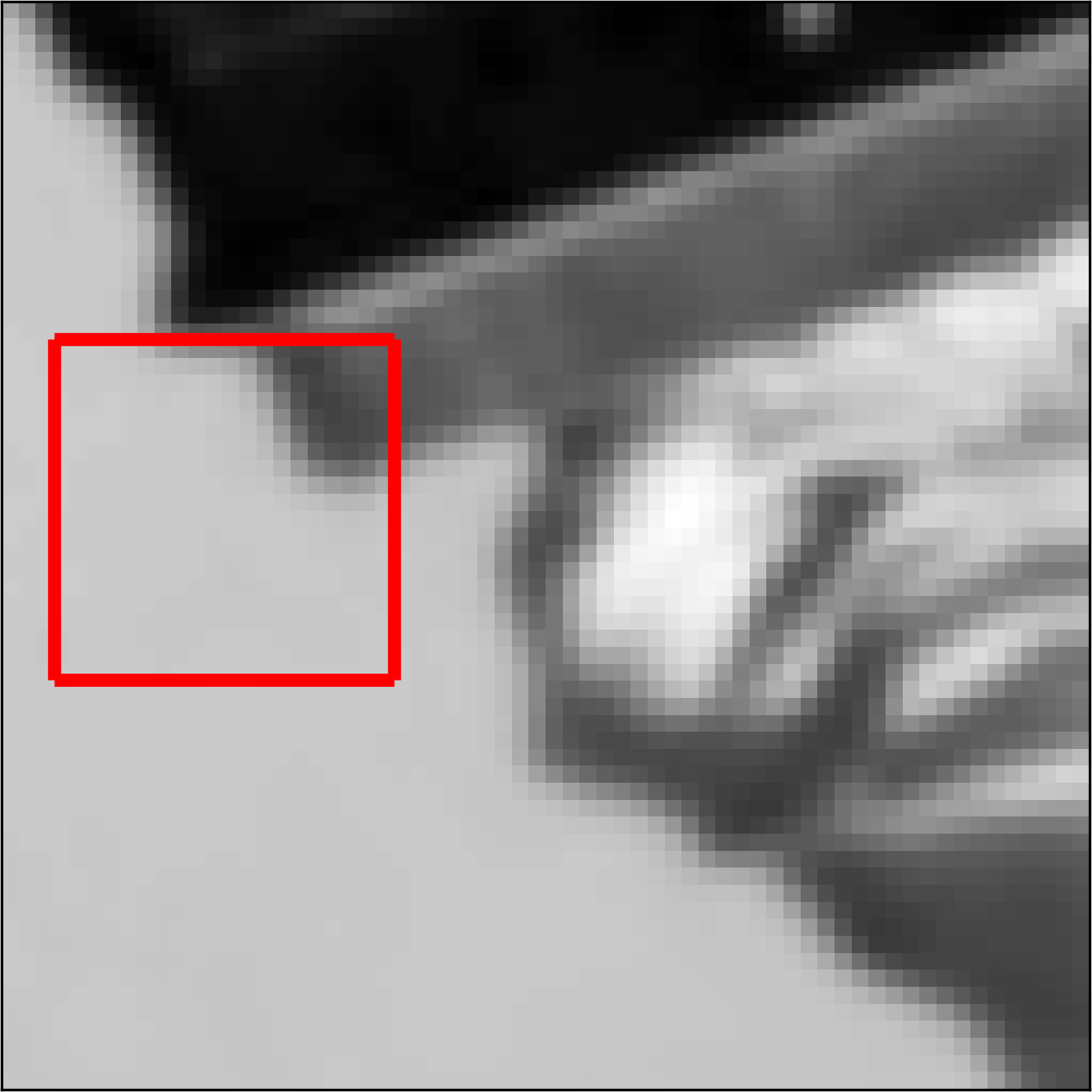}
  & \includegraphics[height=46pt]{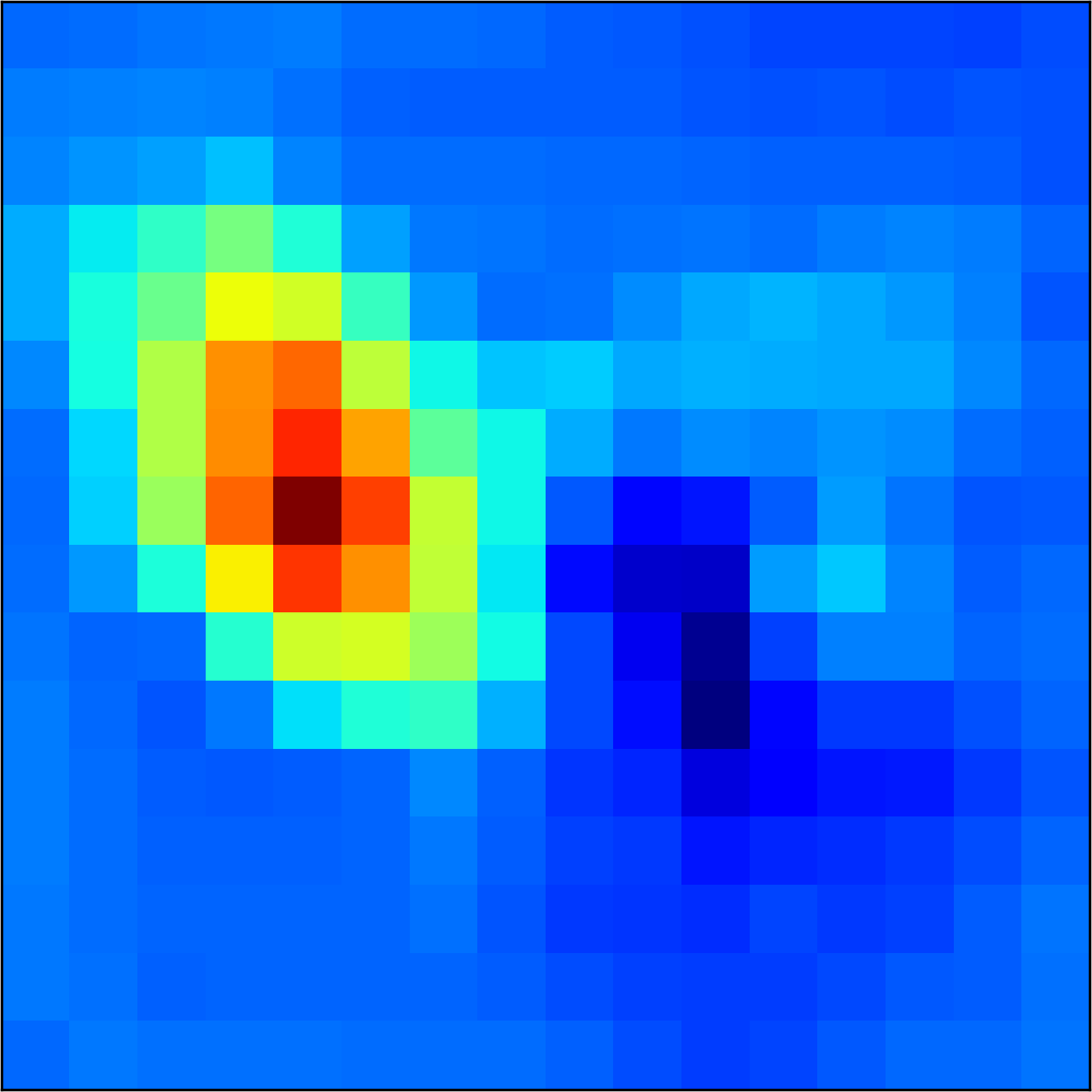}
  & \includegraphics[height=46pt]{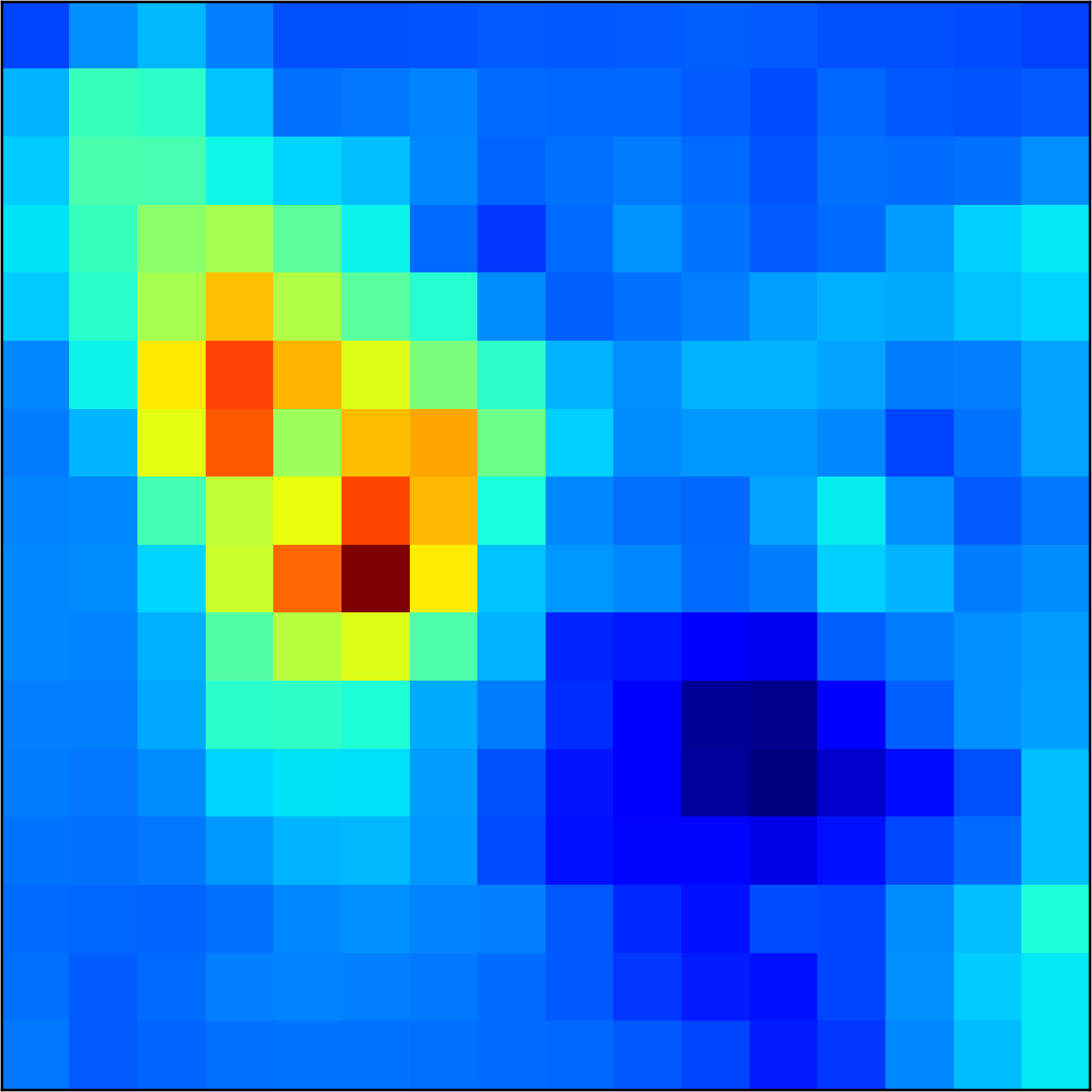}
  & \includegraphics[height=46pt]{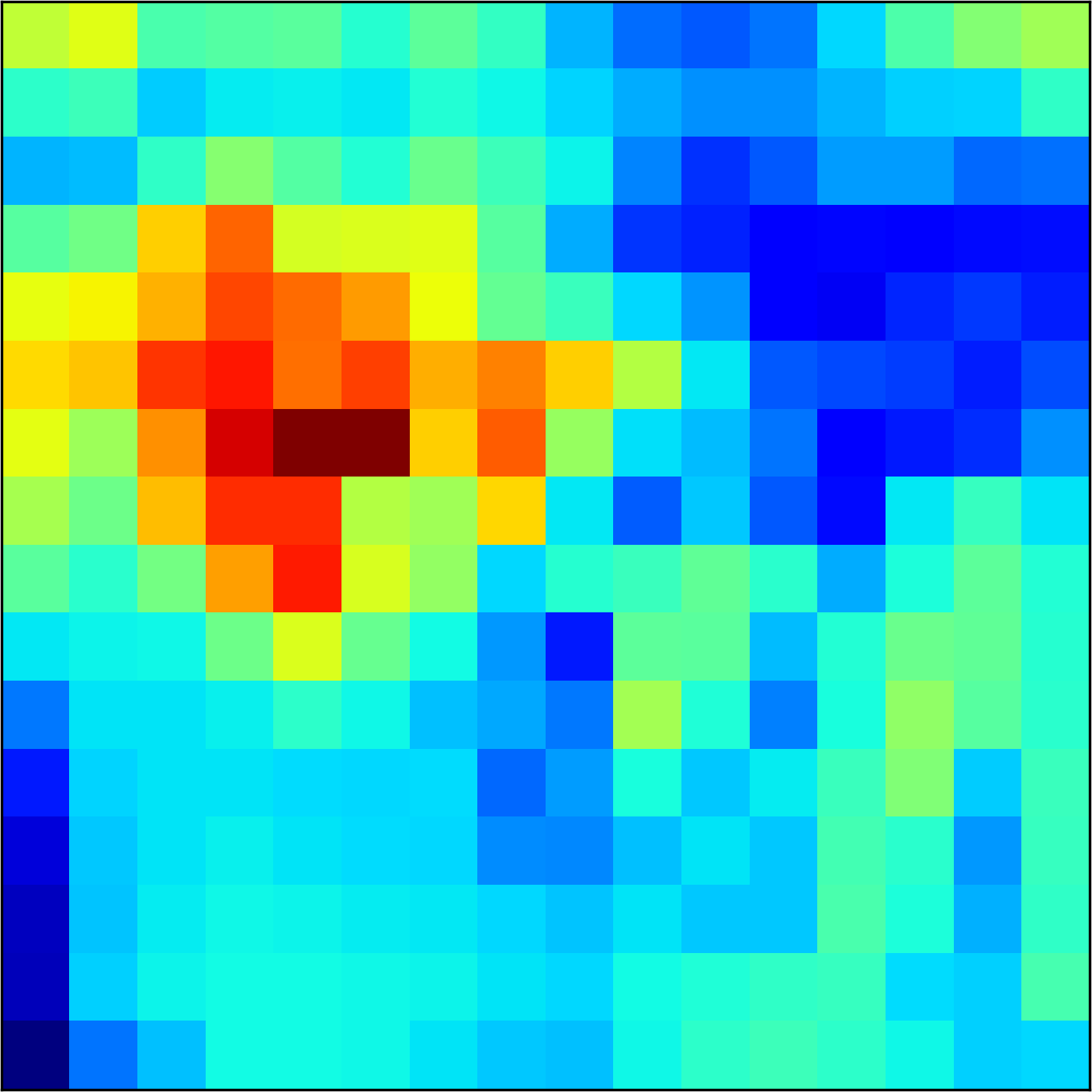}
  & \includegraphics[height=46pt]{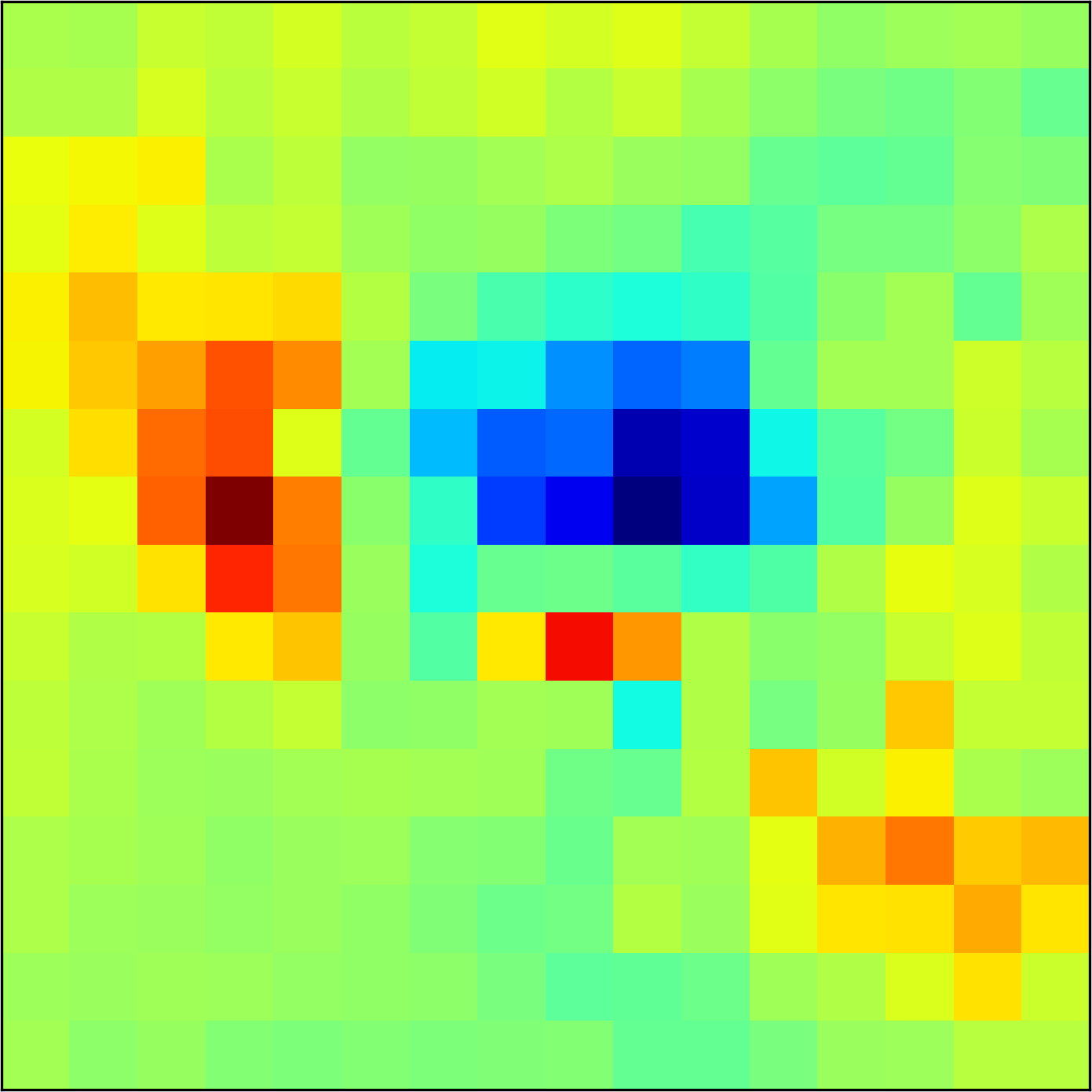} \\
\includegraphics[height=46pt]{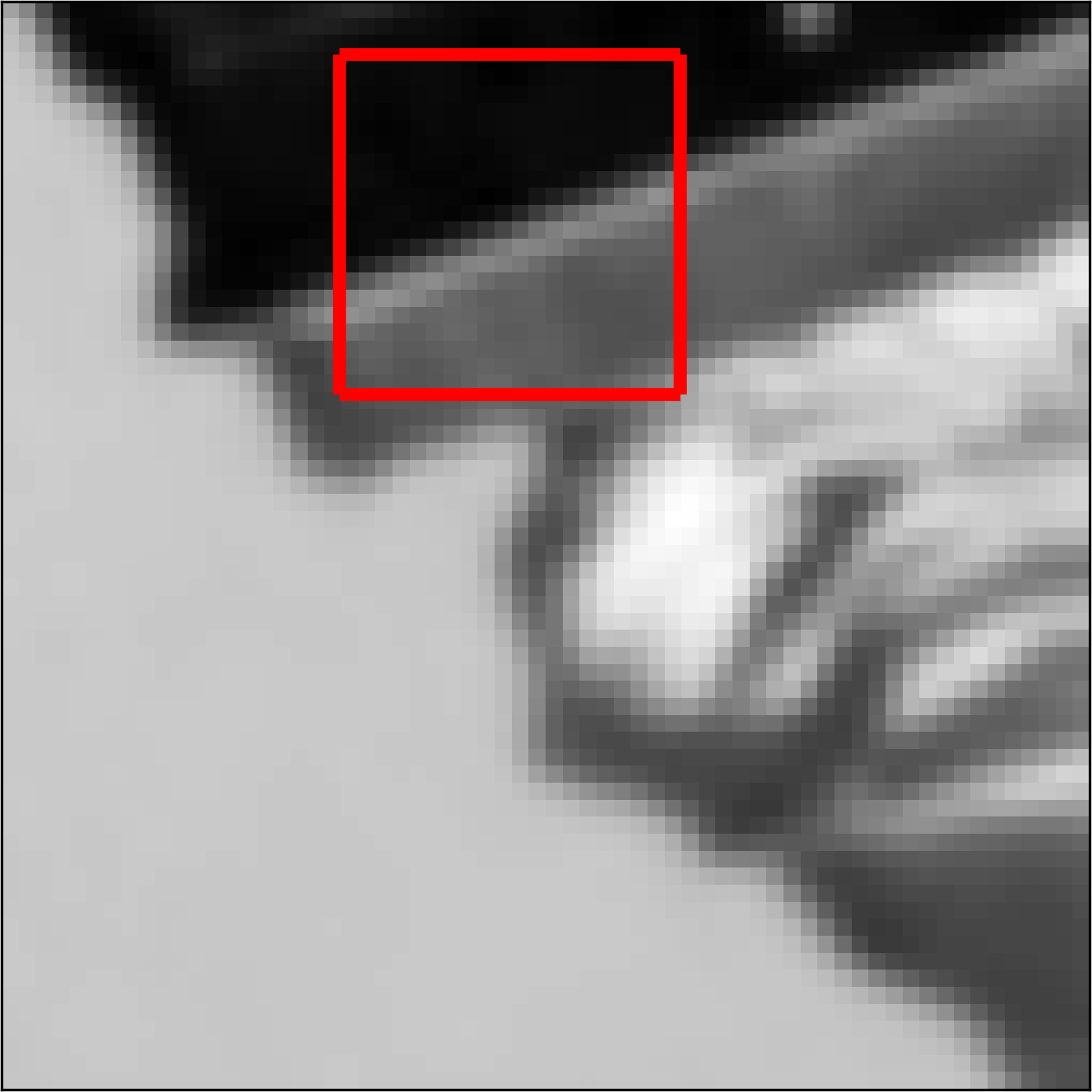}
  & \includegraphics[height=46pt]{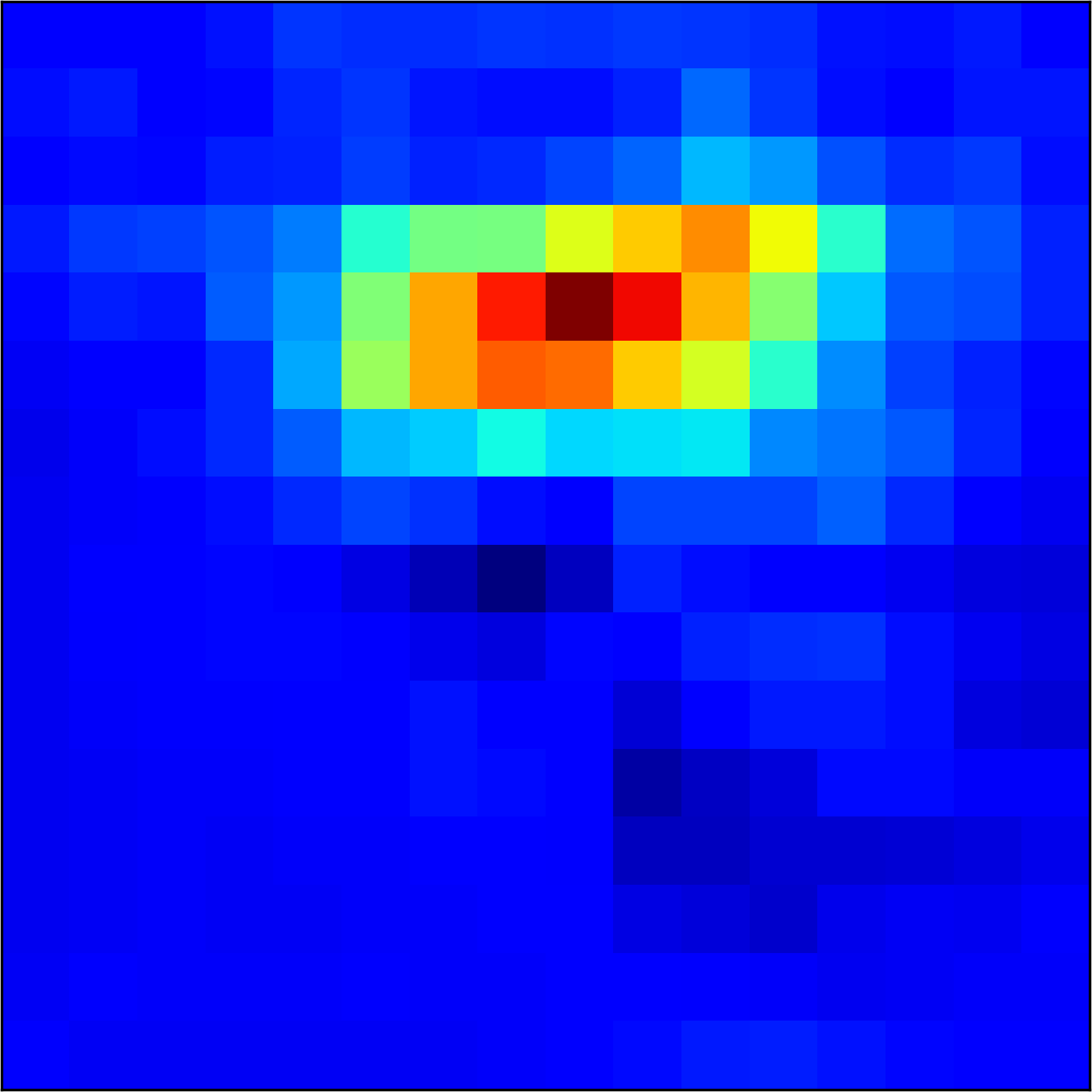}
  & \includegraphics[height=46pt]{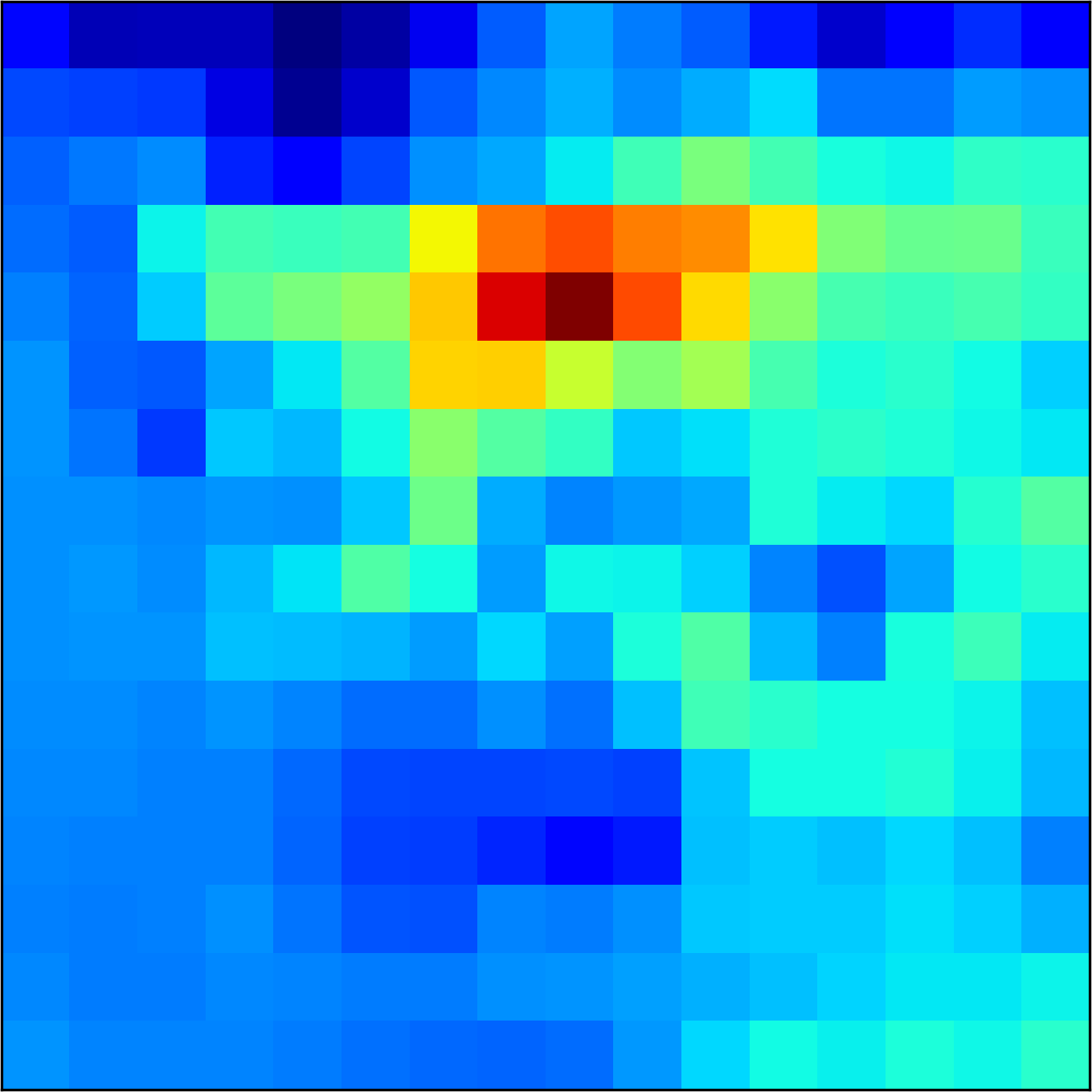}
  & \includegraphics[height=46pt]{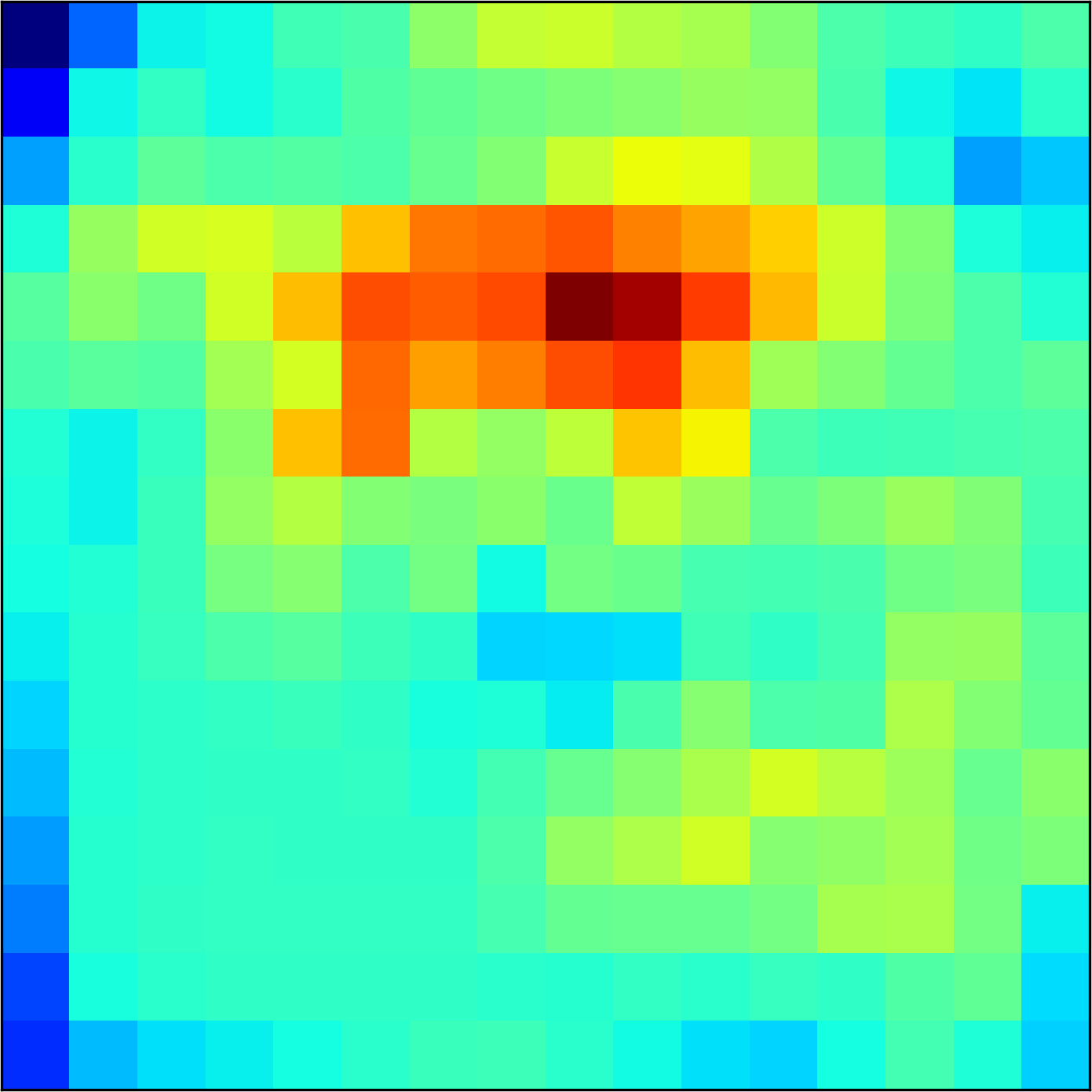}
  & \includegraphics[height=46pt]{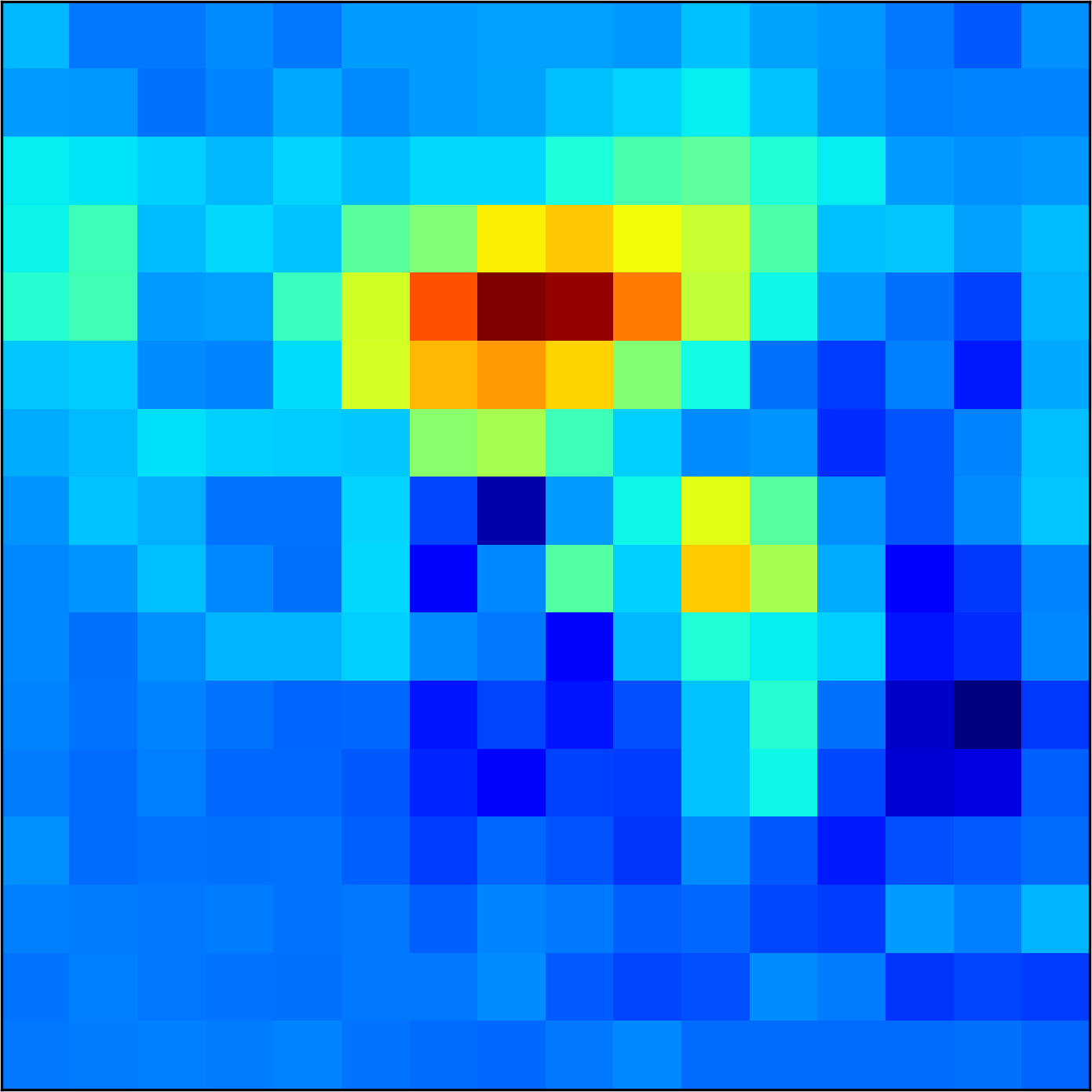} \\
\includegraphics[height=46pt]{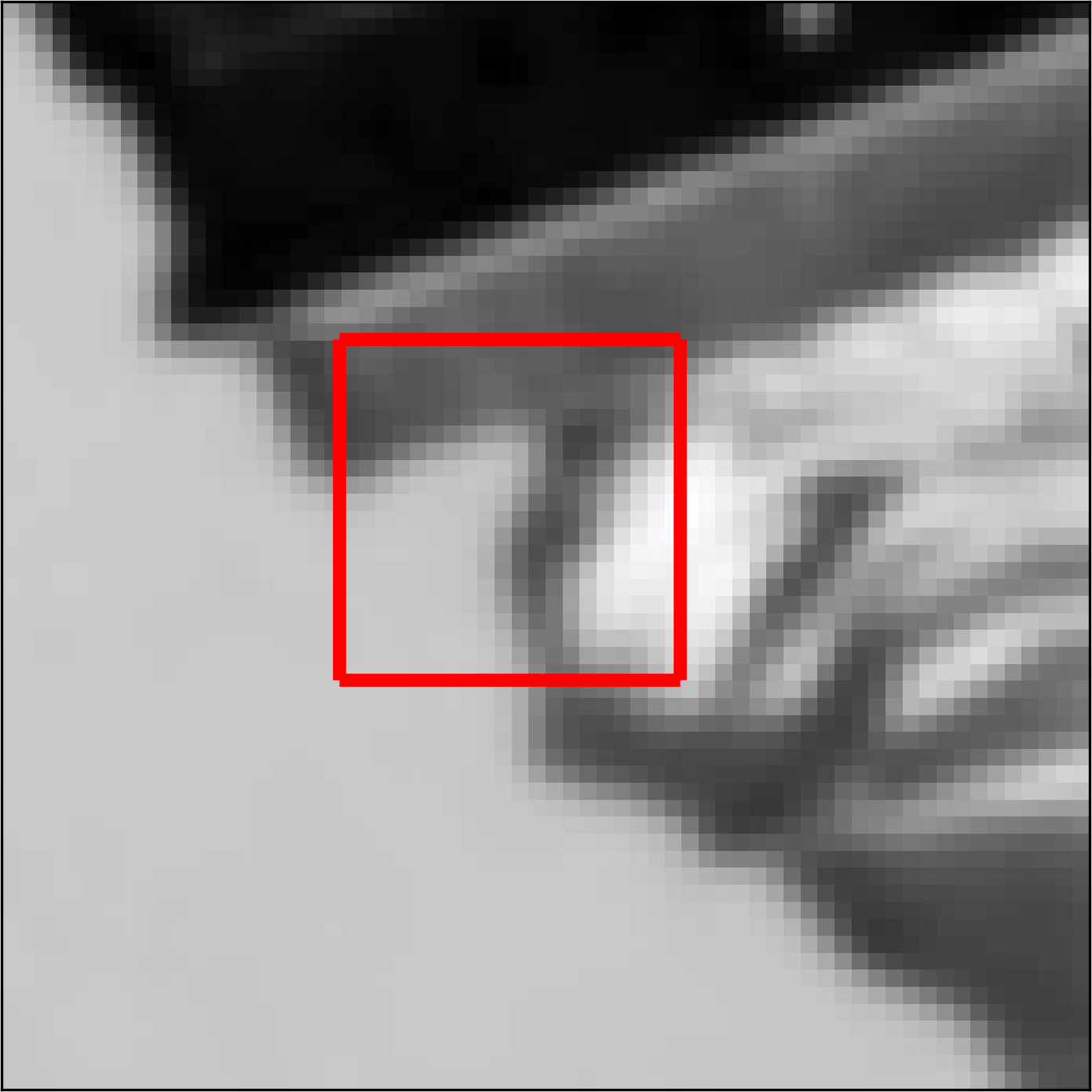}
  & \includegraphics[height=46pt]{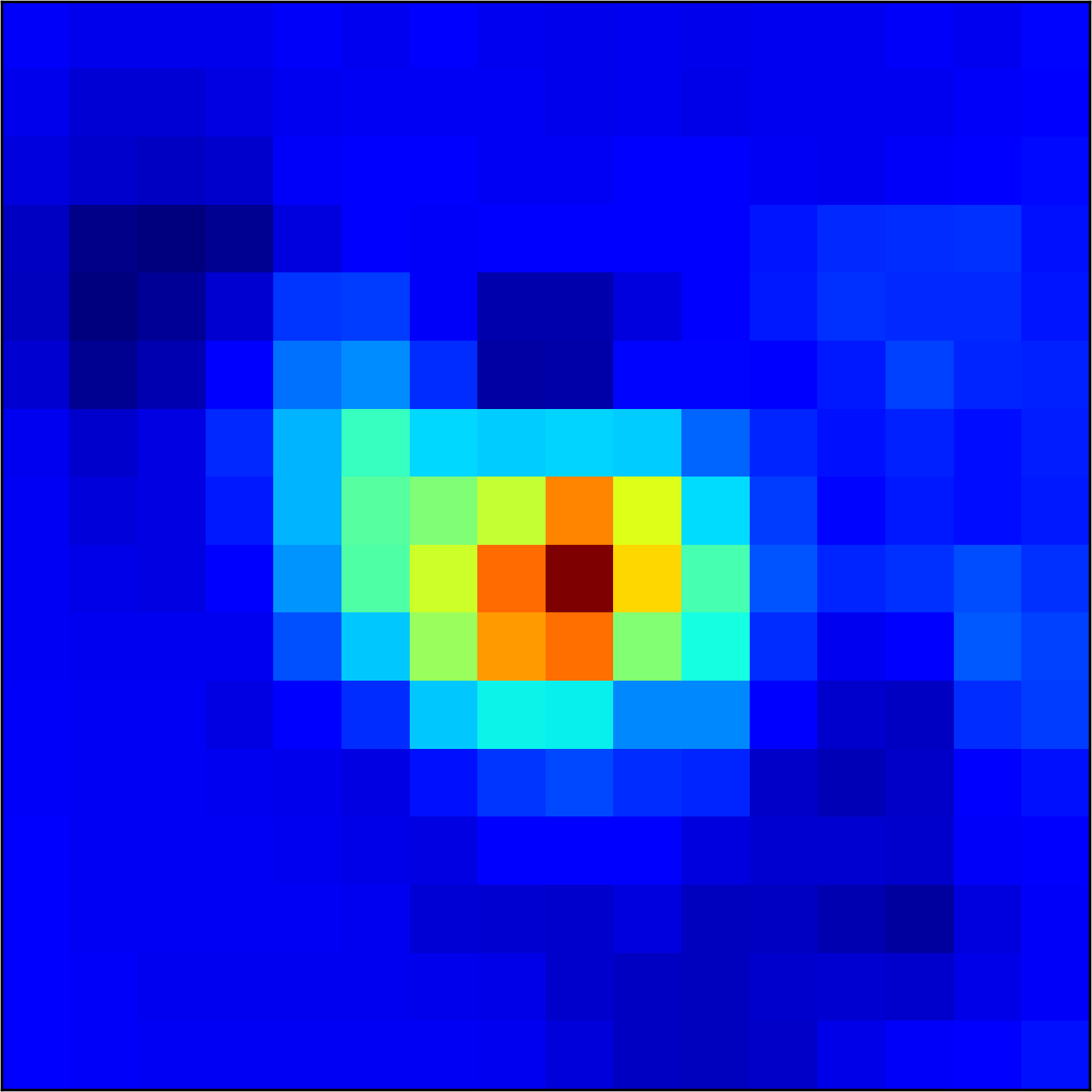}
  & \includegraphics[height=46pt]{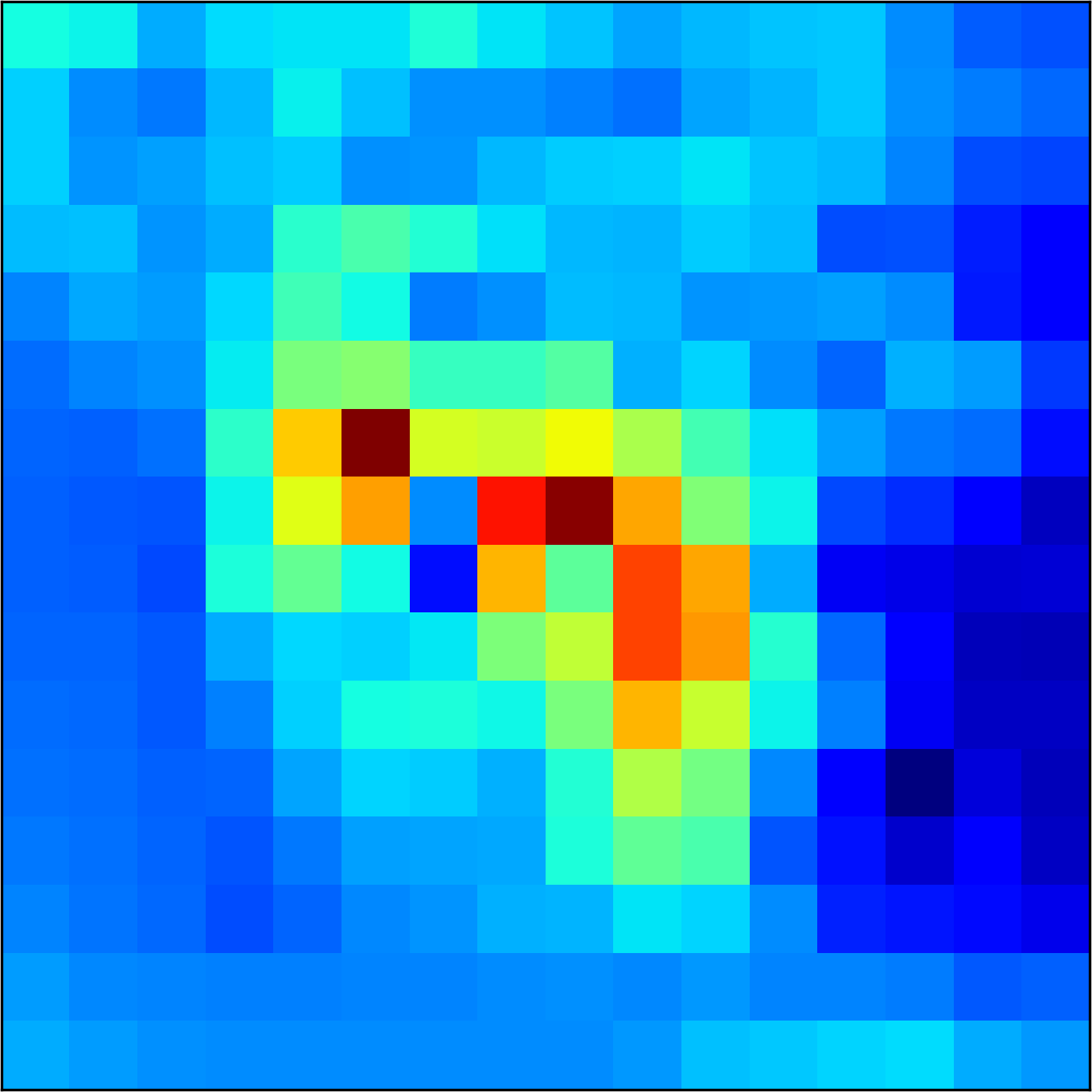}
  & \includegraphics[height=46pt]{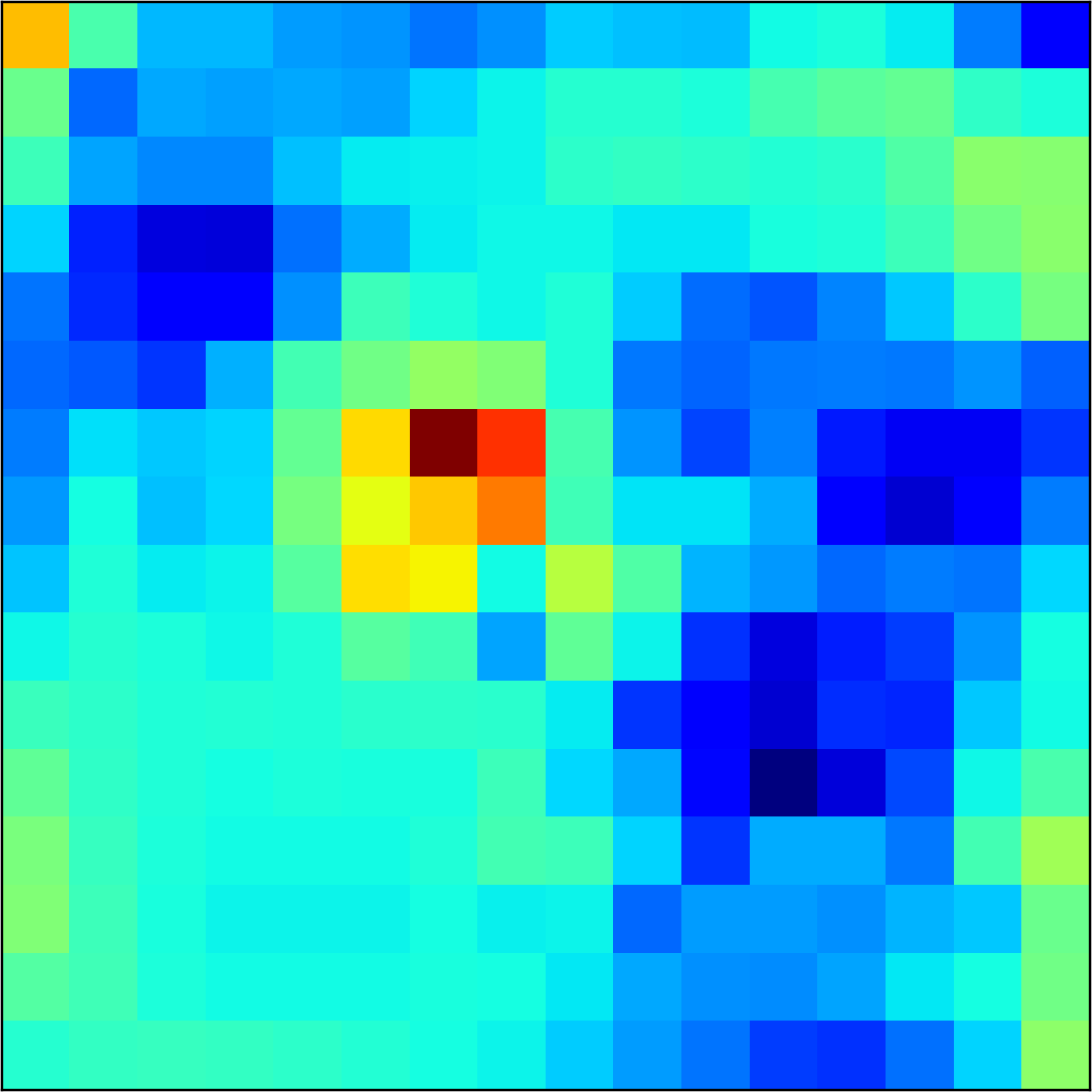}
  & \includegraphics[height=46pt]{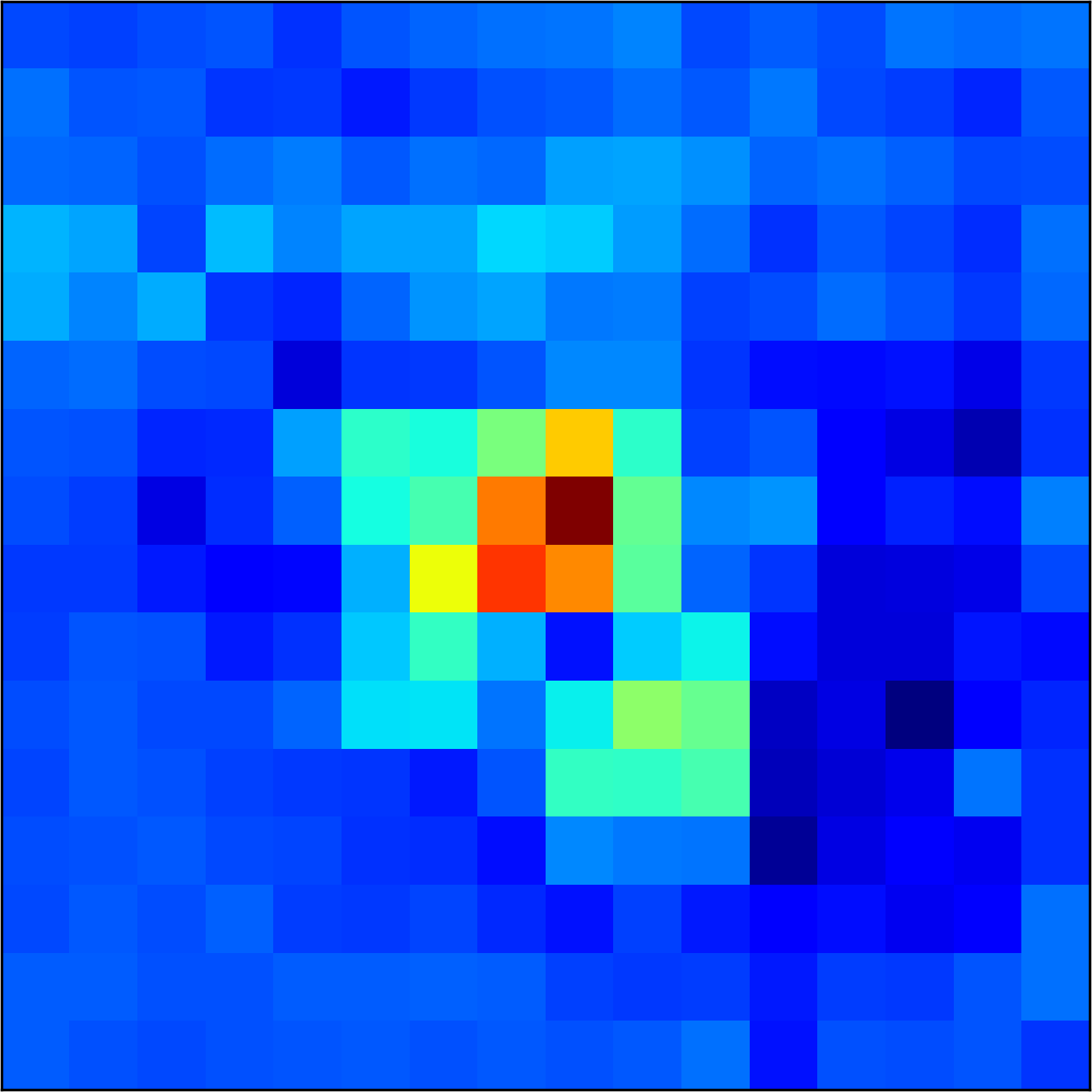} \\
\includegraphics[height=46pt]{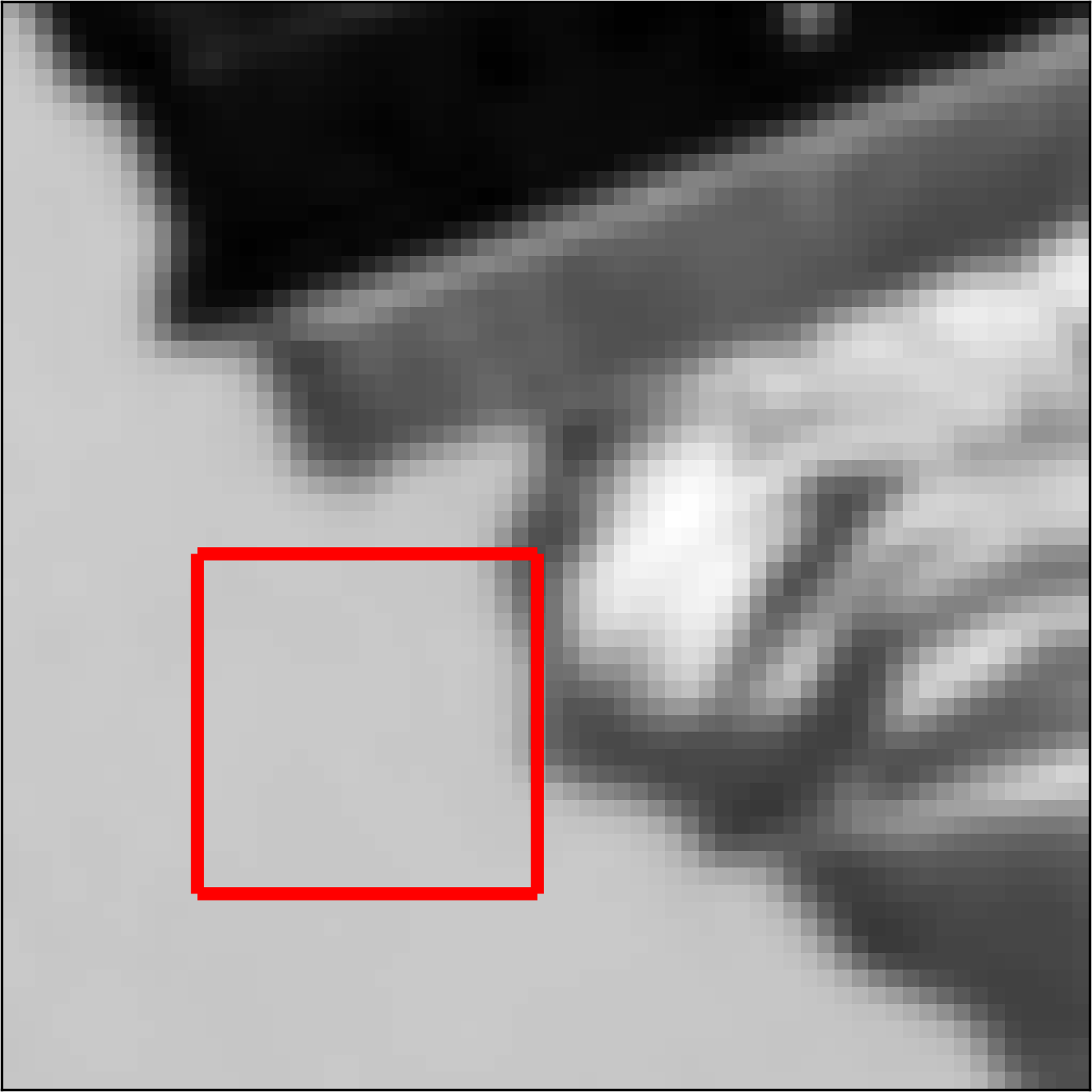}
  & \includegraphics[height=46pt]{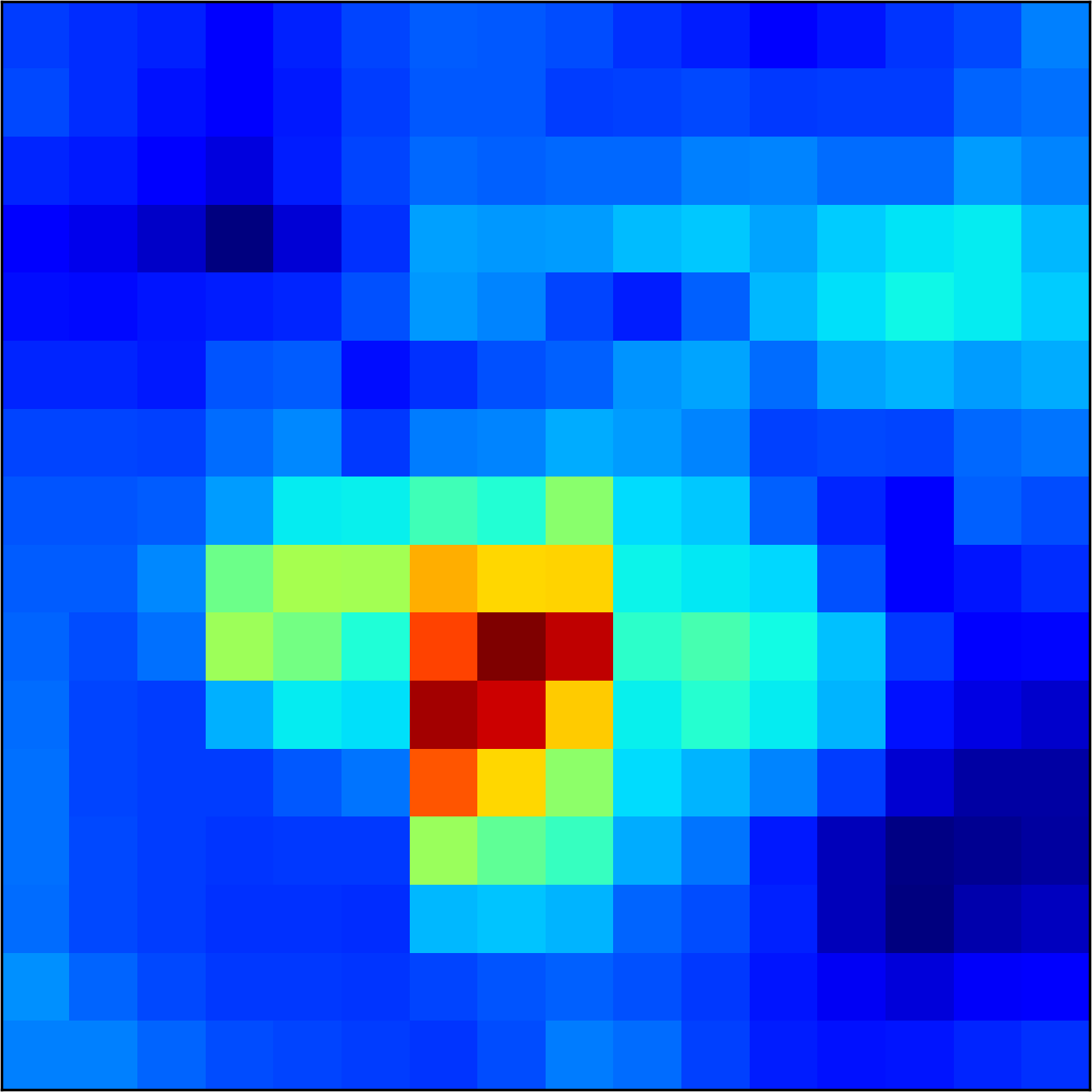}
  & \includegraphics[height=46pt]{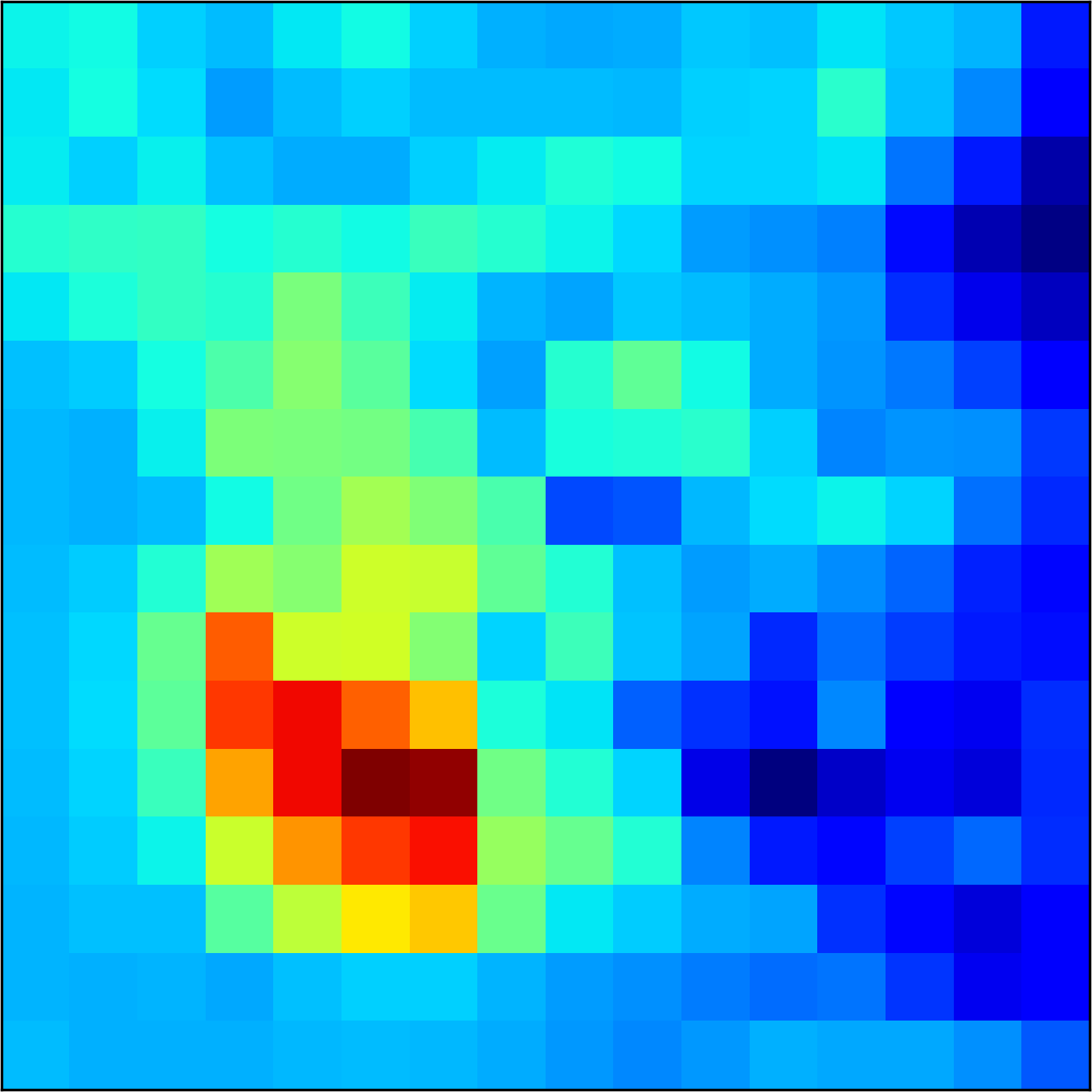}
  & \includegraphics[height=46pt]{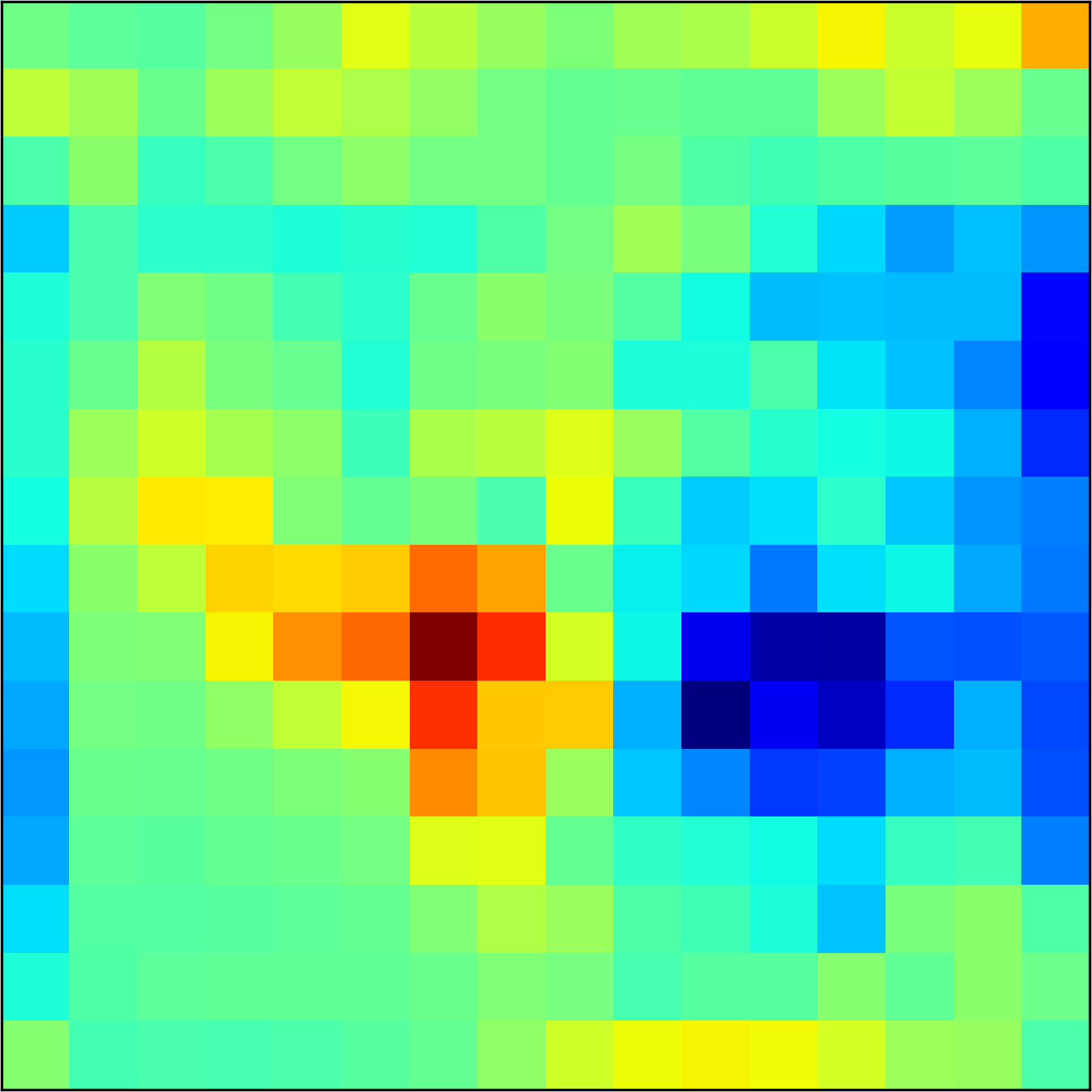}
  & \includegraphics[height=46pt]{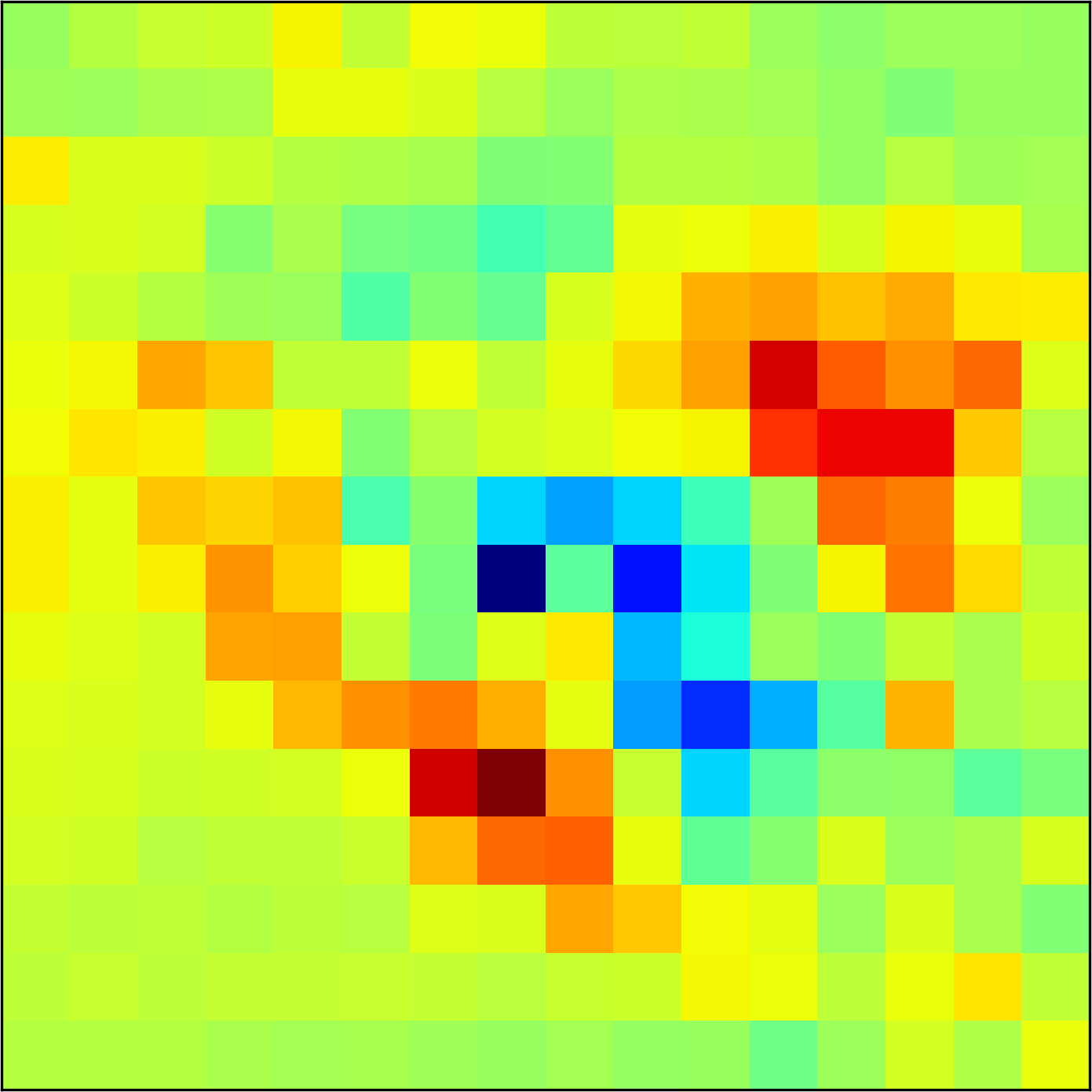} \\
  RF of $p$ & $\vpsi^{xy}$ & $\vpsi^{\rho\theta}$ & $\vpsi^{c}$ & HardNet+\\
\end{tabular}

%% file: tables/pt.tex
\centering
\setlength\extrarowheight{6pt}
\begin{tabular}{lrccccccc}
\toprule
Test & & & \multicolumn{2}{c}{Liberty} & \multicolumn{2}{c}{Notredame} & \multicolumn{2}{c}{Yosemite} \\ \cmidrule(lr){1-1}\cmidrule(lr){4-5}\cmidrule(lr){6-7}\cmidrule(lr){8-9}
Train & \# Parameters & Mean & No& Yo& Li & Yo & Li & No  \\\midrule
\midrule[\heavyrulewidth]
  HardNet+ $\dagger$     &  1,334,560   & \quad  1.51 {\color{white}\scriptsize \pmm 0.00}  \quad  & 1.49{\color{white}\scriptsize \pmm 0.00}           & 2.51{\color{white}\scriptsize \pmm 0.00}           & 0.53{\color{white}\scriptsize \pmm 0.00}           & 0.78{\color{white}\scriptsize \pmm 0.00}           & 1.96{\color{white}\scriptsize \pmm 0.00}           & 1.84{\color{white}\scriptsize \pmm 0.00} \\
  HardNet+ 	     &  1,334,560   & \quad  1.43 {\scriptsize \pmm 0.02}  \quad  & 1.25 {\scriptsize \pmm 0.03} & 2.35 {\scriptsize \pmm 0.03} & 0.48 {\scriptsize \pmm 0.01} & 0.74 {\scriptsize \pmm 0.02} & 2.15 {\scriptsize \pmm 0.01} & 1.61 {\scriptsize \pmm 0.10} \\
  \indepa                &    867,008   & \quad  1.53 {\scriptsize \pmm 0.03}  \quad  & 1.27 {\scriptsize \pmm 0.03} & 2.31 {\scriptsize \pmm 0.08} & 0.48 {\scriptsize \pmm 0.02} & 0.82 {\scriptsize \pmm 0.05} & 2.58 {\scriptsize \pmm 0.08} & 1.72 {\scriptsize \pmm 0.09} \\
  \indepb                &  1,391,296   & \quad  1.36 {\scriptsize \pmm 0.01}  \quad  & 1.14 {\scriptsize \pmm 0.03} & 2.16 {\scriptsize \pmm 0.10} & 0.42 {\scriptsize \pmm 0.01} & 0.73 {\scriptsize \pmm 0.02} & 2.18 {\scriptsize \pmm 0.07} & 1.51 {\scriptsize \pmm 0.12} \\
\bottomrule
\end{tabular}

%% file: tables/hp_internal.tex
\centering \footnotesize
\setlength{\figH}{3.8cm}
\setlength{\figW}{0.32\columnwidth}
\input{tables/internal/verif}
\input{tables/internal/match}
\input{tables/internal/retrv}

%% file: tables/hp_stateofart.tex
\centering \footnotesize
\setlength{\figH}{4.6cm}
\setlength{\figW}{0.36\columnwidth}
\input{tables/stateofart/verif}
\input{tables/stateofart/match}
\input{tables/stateofart/retrv}